\documentclass[nohyperref]{article}

\usepackage{microtype}
\usepackage{graphicx}
\usepackage{subfigure}
\usepackage{booktabs} 
\newcommand{\defeq}{\mathrel{\stackrel{\textnormal{\tiny def}}{=}}}

\usepackage{hyperref}

\newcommand*\circled[1]{\textcircled{\raisebox{-0.9pt}{#1}}}

\usepackage[accepted]{icml2022}

\usepackage{amsmath}
\usepackage{amssymb}
\usepackage{mathtools}
\usepackage{amsthm}
\newcommand{\classificationscore}{\textsc{CA}}

\usepackage[capitalize,noabbrev]{cleveref}

\theoremstyle{plain}

\theoremstyle{definition}

\theoremstyle{remark}

\usepackage[textsize=tiny]{todonotes}

\icmltitlerunning{}

\begin{document}

\twocolumn[
\icmltitle{Describing Differences between Text Distributions with Natural Language}




\begin{icmlauthorlist}
\icmlauthor{Ruiqi Zhong}{ucb}
\icmlauthor{Charlie Snell}{ucb}
\icmlauthor{Dan Klein}{ucb}
\icmlauthor{Jacob Steinhardt}{ucb}
\end{icmlauthorlist}

\icmlaffiliation{ucb}{Computer Science Division, University of California, Berkeley}

\icmlcorrespondingauthor{Ruiqi Zhong}{ruiqi-zhong@berkeley.edu}

\icmlkeywords{Machine Learning, ICML}
\vskip 0.3in
]\printAffiliationsAndNotice{}




\begin{abstract}
How do two \textit{distributions} of text differ?
Humans are slow at answering this, since discovering patterns might require tediously reading through hundreds of samples.
We propose to automatically describe the differences by ``learning a natural language hypothesis'':
given two distributions $D_{0}$ and $D_{1}$, we search for a description that is more often true for $D_{1}$, e.g., ``\textit{is military-related.}''
To tackle this problem, we fine-tune GPT-3 to propose descriptions with the prompt: ``[samples of $D_{0}$] + [samples of $D_{1}$] + \textit{the difference between them is \underline{\space\space\space\space}}.''
We then re-rank the descriptions by checking how often they hold on a larger set of samples with a learned verifier.
On a benchmark of 54 real-world binary classification tasks, while GPT-3 Curie (13B) only generates a description similar to human annotation 7\% of the time, the performance reaches 61\% with fine-tuning and re-ranking, and our best system using GPT-3 Davinci (175B) reaches 76\%.
We apply our system to describe distribution shifts, debug dataset shortcuts, summarize unknown tasks, and label text clusters, and present analyses based on automatically generated descriptions.

\end{abstract}

\section{Introduction}
What inputs trigger a neuron in my deep learning model?
How are the train and test distributions different for my application?
How did public opinions on Twitter change from last year to this year?
These questions have significant scientific, economic, and social consequences.
However, discovering new patterns sometimes requires scanning over thousands of examples, intractable for humans.
An automated solution would be far more scalable.

\begin{figure}[t!]
    \centering
    \includegraphics[width=\columnwidth]{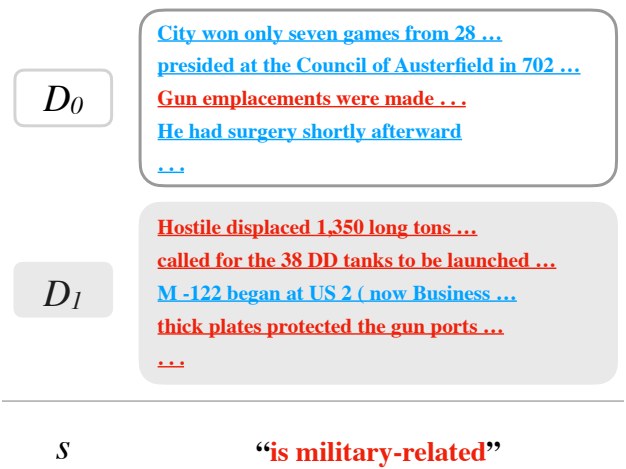}
    \caption{
    Given two distributions (top), our system automatically discovers their differences and describes them with natural language (bottom).
    Grey/white background represents $D_{0}$/$D_{1}$ and red/blue represents whether a sample matches the description $s$.
    }
    \label{fig:fig1}
\end{figure}

To address this, we develop a method to discover the differences between two distributions and describe them with natural language. 
We reduce the above questions to ``learning a natural language hypothesis'' (\Cref{sec:definition}): 
given two text distributions $D_{0}$ and $D_{1}$, we search for a natural language hypothesis $s$ that is more often true for samples from $D_{1}$ than samples from $D_{0}$.
For instance:
\begin{itemize}
    \item We can describe what triggers an artificial neuron by setting $D_{1}$ to be inputs that trigger it and $D_{0}$ for other inputs. $s$ could be ``\textit{is military-related}'' (\Cref{fig:fig1}).
    \item We can describe the differences between the train and test distributions by setting them to be $D_{0}$ and $D_{1}$. 
    A possible $s$ would be ``\textit{is longer in sentence length}.''
    \item We can describe how public opinions shifted by setting $D_{0}$/$D_{1}$ to be the opinions from last year/this year. $s$ could be ``\textit{is optimistic about the pandemic}.''
\end{itemize}

\begin{figure*}[t!]
    \centering
    \includegraphics[width=\textwidth]{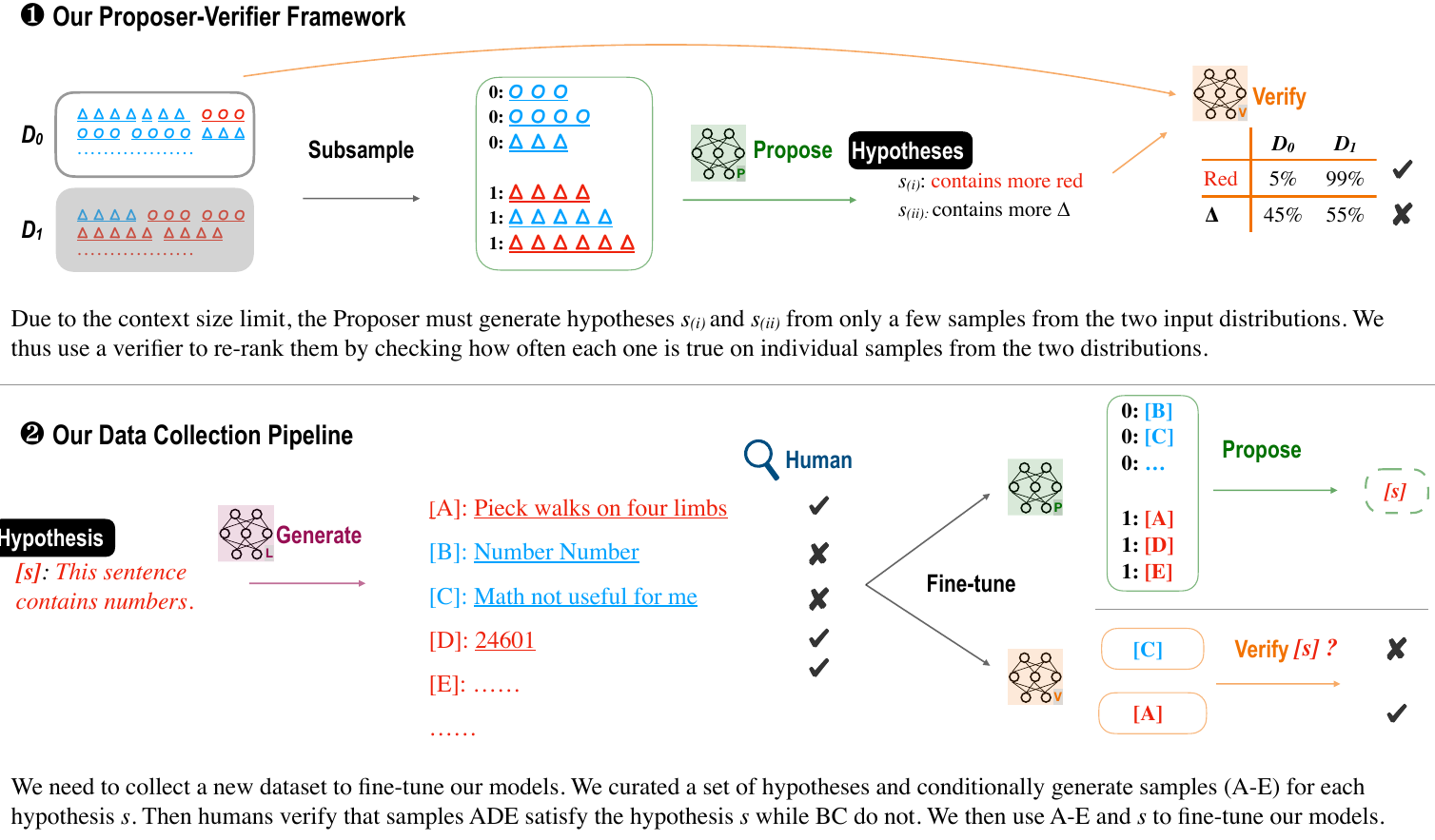}
    \caption{
    Our architectural framework (\textbf{top}) and data collection pipeline (\textbf{bottom}). \Cref{sec:method} describes them in detail.}
    \label{fig:main}
\end{figure*}

\begin{figure}[t!]
    \centering
    \includegraphics[width=\columnwidth]{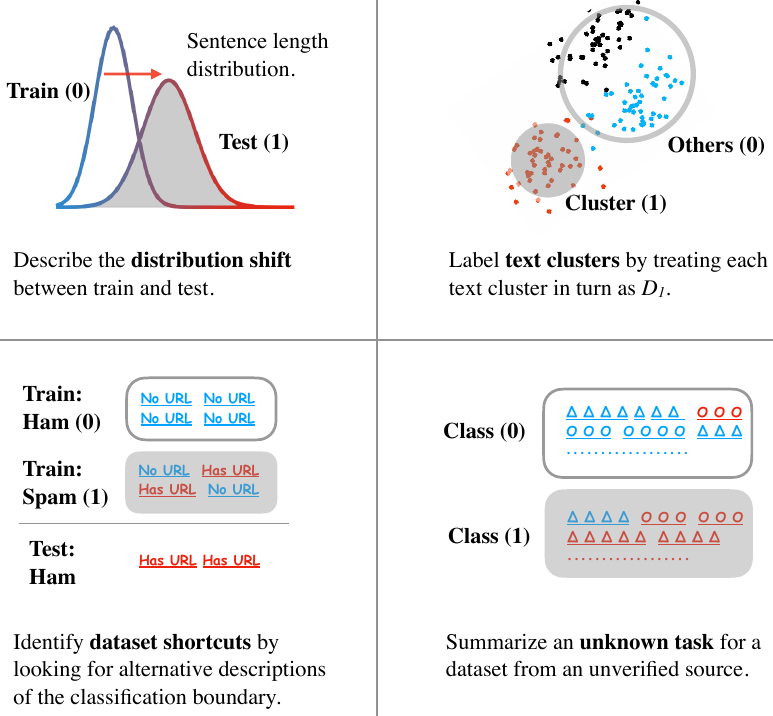}
    \caption{We reduce a wide range of applications to learning a natural language hypothesis and present our analyses in \Cref{sec:application}.}
    \label{fig:application}
\end{figure}

\begin{figure*}[t!]
    \centering
    \includegraphics[width=0.9\textwidth]{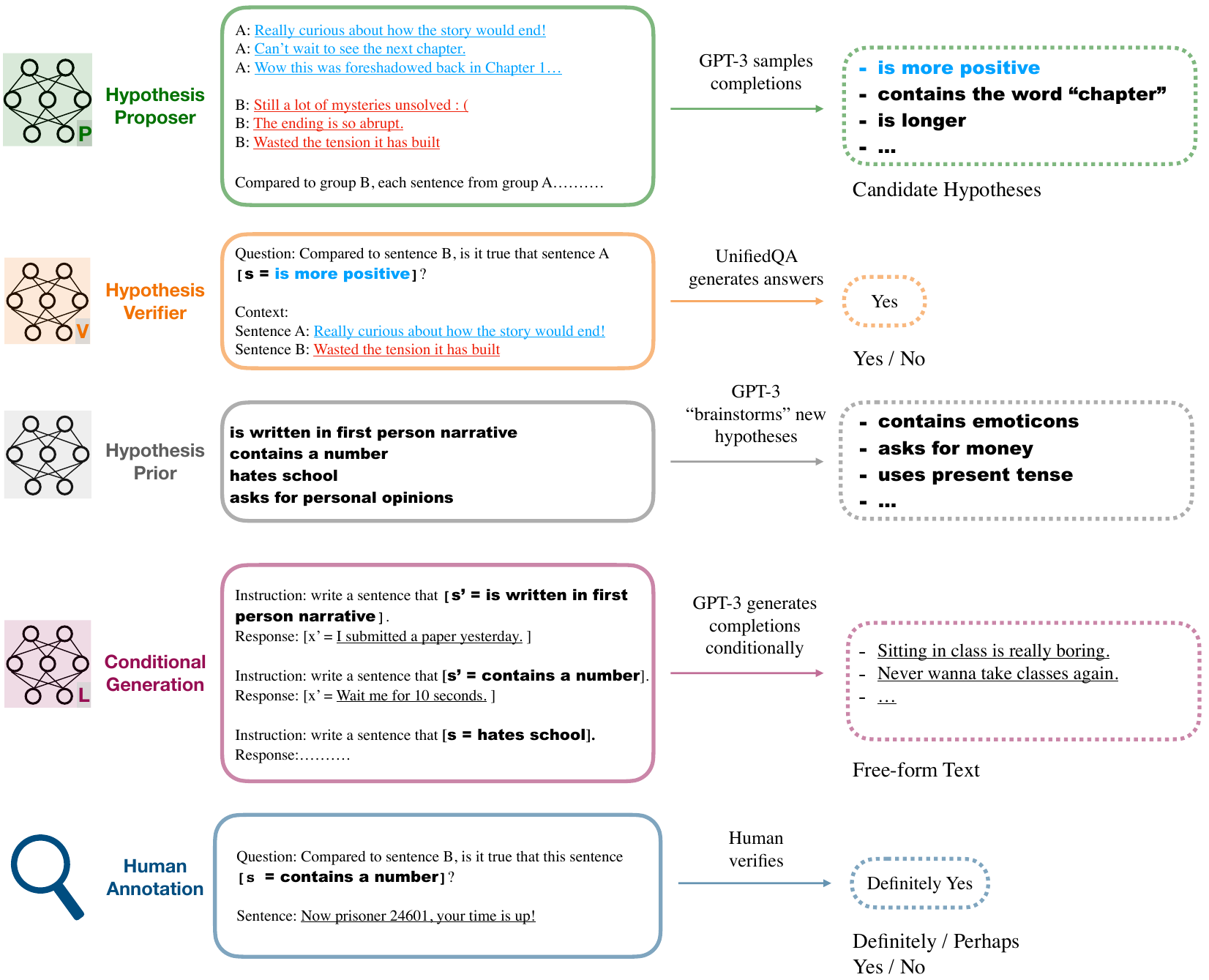}
    \caption{The prompt template for all components in our system. 
    All text datapoints $x$ are underlined and hypotheses $s$ bolded.}
    \label{fig:prompt-format}
\end{figure*}

We develop a new method to learn a natural language hypothesis.
We first prompt GPT-3 Davinci (175B) \cite{brown2020language} with samples from each distribution and ask it to propose candidate hypotheses $s$ (\Cref{sec:proposer}).
However, since GPT-3 has a limited context size, this prompt can only contain a few samples rather than the whole distributions.
Therefore, we re-rank the candidates with a verifier that checks how often they hold on a larger set of samples (\Cref{sec:verifier}).
We visualize our framework at the top of \Cref{fig:main} and the prompts at the top of \Cref{fig:prompt-format}.

Since GPT-3 is not optimized to propose hypotheses, we can improve it through fine-tuning.
However, no corpus exists for this task yet. 
Therefore, we developed a new data collection pipeline (\Cref{sec:collect}) with three stages:
1) we curated a list of hypotheses $s$, 2) we asked GPT-3 to generate samples that satisfy $s$, and 3) we asked annotators to judge whether they indeed satisfy $s$.
Then we fine-tuned the proposer to predict $s$ based on samples that satisfy $s$ and samples that do not (\Cref{sec:fine-tuning}).
We visualize our data collection and fine-tuning method at the bottom of \Cref{fig:main}.

We benchmark our system on 54 real-world binary classification datasets \cite{zhong-etal-2021-adapting-language}, each annotated with natural language descriptions for the positive class.
For each binary task, we treat the positive/negative class inputs as $D_{1}/D_{0}$ and compare the top-5 descriptions by our system to the human annotation.
While the descriptions by GPT-3 Curie (13B) are similar to the annotations only 7\% of the time, the performance reaches 61\% with fine-tuning and verifier re-ranking, and our best system using GPT-3 Davinci (175B) reaches 76\% (\Cref{sec:benchmark}).

We then check whether the intended uses of existing classification datasets agree with the descriptions by our system (\Cref{sec:application}).
Our system correctly recognizes that the subjectivity analysis (SUBJ) dataset \cite{pang-lee-2004-sentimental} was constructed by contrasting movie reviews with plot summaries; however, many recent papers \cite{bragg2021flex, zhong-etal-2021-adapting-language, gao-etal-2021-making, min2021noisy} were unaware of this fact and used SUBJ for zero/few-shot subjectivity classification.
Our system also recognizes several dataset shortcuts. 
For example, it rediscovered that negations, such as the use of ``not/never'', is spuriously correlated with the contradiction class in MNLI \cite{gururangan-etal-2018-annotation};
for another example, models trained on the SMS Spam classification dataset \cite{gomez2006content} always consider hyperlinks to be spam.
Our system can also describe distribution shifts and text clusters (\Cref{sec:application}), and \Cref{fig:application} visualizes all our applications.
We conclude with future applications in other modalities (e.g., vision) and research fields (e.g., social science) in \Cref{sec:disussion}.\footnote{\cref{appendix:code} discusses details about code and data release.}

\section{Learning a Natural Language Hypothesis} \label{sec:definition}

Let $\mathcal{X}$ be the set of all text inputs.
A natural language hypothesis $h$ is parameterized by a natural language string $s$ and is a mapping from two inputs to a boolean:
\begin{equation}
    h_{s}: \mathcal{X} \times \mathcal{X} \rightarrow \{0, 1\},
\end{equation}
where $h_{s}(x_{1}, x_{0}) = 1$ means $x_{1}$ is more $s$ than $x_{0}$.
For example, if $s$ is ``\textit{is longer in sentence length},'' then $h_{s}(x_{1}, x_{0}) = 1$ means $x_{1}$ is longer than $x_{0}$.
The semantics of $h_{s}$ is defined as
\begin{equation} \label{eq:human-thinks}
    h_{s}(x_{1}, x_{0}) \defeq \mathbf{1}[\text{humans consider }x_{1} \text{ more }s \text{ than } x_{0}],
\end{equation}

which our paper operationalizes by taking majority vote among crowdworkers.\footnote{More broadly, however, there is no canonical method to interpret natural language. See \Cref{sec:disussion} for more discussion.}
We call both $s$ and $h_{s}$ ``hypotheses'' but write $s$ when using it as a string and $h_{s}$ as a function.

Let $D_{0}$ and $D_{1}$ be two distributions over $\mathcal{X}$, and $\mathcal{H}$ be the space of all valid natural language hypotheses.
We search for $h$ in $\mathcal{H}$ to maximize its ``classification accuracy'' $\classificationscore$,
\begin{equation} \label{eq:gold-loss}
    \classificationscore(h) \defeq \mathbb{E}_{x_{0}\sim D_{0}, x_{1}\sim D_{1}}[h(x_{1}, x_{0}))].
\end{equation}

Intuitively, given two random samples from each distribution $x_{0} \sim D_{0}$ and $x_{1} \sim D_{1}$, $h$ should classify where each $x$ comes from as accurately as possible.
Therefore, our task falls under the standard formulation of statistical machine learning, where we learn a hypothesis $h$ by optimizing a statistical objective ($\classificationscore$) over a hypothesis space $\mathcal{H}$. 

Compared to traditional statistical learning, learning a natural language hypothesis poses two new challenges.

\noindent\textbf{Search.} Searching in a discrete string space is hard.
\Cref{sec:proposer} addresses this by proposing $h_{s}$ with a neural network based on samples from $D_{0}$ and $D_{1}$.

\noindent\textbf{Verify.} Computing $h_{s}(x_{1}, x_{0})$ requires human annotations, which can be expensive.
\Cref{sec:verifier} addresses this by approximating human responses with a neural network.

\section{Method} \label{sec:method}
We prompt GPT-3 to propose hypotheses based on a small set of samples (\Cref{sec:proposer}) and use UnifiedQA to verify each hypothesis on a larger set of samples (\Cref{sec:verifier}).
Then, we design a data collection pipeline (\Cref{sec:collect}) to further fine-tune the proposer and the verifier  (\Cref{sec:fine-tuning}).
Our methods can be visualized in \Cref{fig:main}.

\subsection{Hypothesis Proposer} \label{sec:proposer}
Our goal is to generate a list of plausible hypotheses based on samples from $D_{0}$ and $D_{1}$.
We do so by prompting GPT-3, a language model that can generate textual completions based on a prompt. 
We construct a ``proposer prompt'' by concatenating several samples from $D_{1}$, several from $D_{0}$, and the instruction ``\textit{Compared to group 0, each sentence from group 1 \underline{\space\space\space\space}}'' (\Cref{fig:prompt-format}, the 1$^{\text{st}}$ row).
Since GPT-3 has a context size limit of 2048, we select 5 samples $x$ from each distribution.

Without controlled decoding, a typical prompt completion would be ``\textit{is more positive, while sentences from group 0 are ungrammatical.}''
However, such a completion is undesirable, since 1) the verifier now needs to check two statements at the same time, namely, whether samples from $D_{1}$ are positive and samples from $D_{0}$ are ungrammatical, and 2) the second half of the completion describes a population-level property of ``group 0'', while our verifier only checks hypotheses on individual $x$. 
To produce a single hypothesis about individual $x$, we forbid GPT-3 to decode tokens like ``\textit{group}'' and terminate the generation with token ``,'' or ``.''.

Additionally, $D_{0}$ and $D_{1}$ might overlap, and even an optimal hypothesis $h^{*}$ cannot fully separate them.
As a result, the proposer prompt might contain samples from $D_{1}$ that do not satisfy $h^{*}$, thus confusing the proposer. 
Therefore, we choose samples that are representative of their differences to prompt GPT-3.
To find those samples, we fine-tune RoBERTa-Large \cite{Liu2019RoBERTaAR} to predict whether each sample comes from $D_{0}$ or $D_{1}$ and keep the top-$p$ percentile samples with the highest confidence.
For the top-5, 20, and 100th percentile, we construct proposer prompts with ten different random sets of samples and generate two completions for each set. 
In total we obtain $3\times 10\times 2 = 60$ hypotheses. 
We re-rank them in the next section.

\subsection{Hypothesis Verifier} \label{sec:verifier}
Ideally, we should re-rank $h_{s}$ based on its classification accuracy $\classificationscore(h_{s})$, defined in \cref{eq:gold-loss}.
However, it involves computing $h_{s}(x_{1}, x_{0})$, which requires expensive human annotations (\Cref{eq:human-thinks}).
We therefore approximate it with a verifier neural network $V$:
\begin{equation}
    \hat{h}_{s}(x_{1}, x_{0}) \defeq \frac{1}{2}(V(s, x_{1}, x_{0}) - V(s, x_{0}, x_{1}) + 1).
\end{equation}
Here $V(s, x_{1}, x_{0}) = 1$ if it predicts that $x_{1}$ is more $s$ than $x_{0}$ ($0$ otherwise); then we subtract the baseline $V(s, x_{0}, x_{1})$ obtained by swapping the position of $x_{0}$ and $x_{1}$, and finally normalize the quantity within $[0, 1]$. 

We implement our verifier with UnifiedQA \cite{khashabi-etal-2020-unifiedqa},
a question answering model based on T5 (11B) \cite{raffel2019exploring}. 
UnifiedQA generates an answer $a$ given a question $q$ and a context $c$.
As shown in the 2$^{\text{nd}}$ row of \Cref{fig:prompt-format}, our context $c$ is a pair of sentences $A$ (sampled from $D_1$) and $B$ (sampled from $D_0$). 
The question $q$ is then ``\textit{Is it true that sentence A \textbf{is more positive}}?'', where in general the bolded part is a hypothesis $s$ generated by the proposer.
%
Then we define $V(s, x_1, x_0) = 1$ if UnifiedQA outputs ``yes'' and $0$ if it outputs ``no''.

We now use $V(s, x_{1}, x_{0})$ to compute $\classificationscore(\hat{h}_{s})$ for each candidate $s$ and re-rank them.
To save computation, we estimate $\classificationscore(\hat{h}_{s})$ with 400 random pairs of $(x_{1}, x_{0})$ rather than using the entire datasets.
Finally, we output the top-5 hypotheses to describe how $D_{1}$ and $D_{0}$ differ. 

\subsection{Collecting Data for Supervision} \label{sec:collect}
Since GPT-3 and UnifiedQA are not specifically trained to propose or verify hypotheses, we can improve them by fine-tuning \cite{zhong-etal-2021-adapting-language}.
However, since no corpus exists yet for these tasks, we need to collect a new dataset to fine-tune our models.

To fine-tune the proposer, we want data where the output is a hypothesis $s$ and the input prompt contains five samples that are more $s$ and five that are less $s$.
To fine-tune the verifier, we want tuples $(s, x_{1}, x_{0})$ where $x_{1}$ is more $s$ than $x_{0}$.
Thus for both cases, we want a set of hypotheses $s$, and for each of them, two groups of samples where one group is more $s$ than the other.
We designed our data collection pipeline accordingly: we curated a set of hypotheses $s$, asked GPT-3 to generate samples that do (not) satisfy $s$, and asked humans to filter out failed generations.

\noindent\textbf{Curating Hypotheses.}
We curated a pool of 302 hypotheses by hand with the help of GPT-3 \cite{brown2020language}. 
Concretely, we started the pool by brainstorming ten hypotheses ourselves; then, we sampled five hypotheses from the pool and prompted GPT-3 with their concatenation, as visualized in the 3$^{\text{rd}}$ row of \Cref{fig:prompt-format}.
Whenever GPT-3 completed the prompt with a hypothesis different from our existing ones, we added it to the pool.

Our curated hypotheses ranges from shallow (``\textit{contains the word ``yay'' at the end of the sentence}'') to topical (``\textit{loves school}'') to more complex social and linguistic cues (``\textit{supports universal healthcare},'' ``\textit{is written in first person}'').
To make later conditional generation and human annotation easier, we removed any comparatives from $s$, e.g., removing the word ``\textit{more}'' from ``\textit{loves school more.}''

\noindent\textbf{Conditional Generation.}
We refer to samples that satisfy $s$ as ``positive'' and others as ``negative''.
For example, given $s=$ ``\textit{loves school}'', a positive sample could be ``\textit{My advisor is really helpful and I learned a lot.}''
Both positive and negative samples are necessary to fine-tune our models.

To generate positive samples, we prompted GPT-3 as visualized in the 4$^{\text{th}}$ row of \Cref{fig:prompt-format}: we curated a set of hypotheses $s'$ and their positive samples $x'$ by hand, concatenated them with the target hypothesis $s$, and asked GPT-3 to generate a sample $x$. 
Sometimes, however, $x$ satisfies $s$ due to trivial word overlaps, e.g., $x=$ ``\textit{I love school}'' satisfies $s=$ ``\textit{loves school}.'' 
We curated a list of forbidden output tokens for each hypothesis $s$ by hand to prevent this.

We created negative samples for $s$ by using positive samples for other hypotheses.
If $s$ is highly specific, e.g., ``\textit{talks about microwaves},'' a random sample is unlikely to satisfy it.
Therefore, we treat the positive samples of any other hypotheses as the negative samples for $s$.
However, for $s$ like ``\textit{uses past tense}'', a random sample can satisfy it with non-trivial probability. 
Therefore, we wrote contrast hypotheses such as ``\textit{uses future tense}'' and used their positive samples as the negative samples for $s$. 
Hence, our pool expanded to 352 hypotheses after including newly written ones, and we asked GPT-3 to generate 15 positive samples for each.

\noindent \textbf{Verifying with Human Annotations.}
Some samples $x$ from the conditional generation step do not actually satisfy the hypothesis $s$.
To filter out samples that fail, for each $(s, x)$ pair, we recruited turkers\footnote{We recruited turkers located in the U.S. with $>98\%$ HIT acceptance rate and paid them \$0.04 per HIT; we estimate our pay rate to be \$18/hrs based on how fast the authors perform this task.} to verify whether $x$ satisfies $s$, as visualized in the 5$^{\text{th}}$ row of \Cref{fig:prompt-format}.
We collected three annotations for each $(s,x)$ pair and determined the ground truth by majority vote.
Finally, for each $s$, if fewer than five $x$'s passed the turker vote, the authors wrote additional examples by hand. 

Thus, for each of the initial 302 hypotheses, we obtained at least five positive and five negative samples for it. We next use these to fine-tune our models.

\subsection{Fine-tuning} \label{sec:fine-tuning}
\noindent \textbf{Proposer.} 
For each of the 302 hypotheses $s$, we finetuned GPT-3 to generate $s$ based on five positive and five negative samples.
We used batch size 20 and a small learning rate of 0.05 to prevent memorizing the target.
We fine-tuned for two epochs, each using a different set of subsamples to construct the prompt.

\noindent \textbf{Verifier.}
Given $s$ and a positive/negative sample $x_{1}/x_{0}$, our verifier should predict that $V(s, x_{1}, x_{0}) = 1$ and $V(s, x_{0}, x_{1}) = 0$.
To create a fine-tuning dataset, we randomly sampled 30 positive-negative pairs of $(x_{1}, x_{0})$ for each $s$.
We fine-tuned UnifiedQA on this dataset for 250 steps with batch size 32 and learning rate 5e-5.
To improve out-of-distribution robustness, we averaged the fine-tuned model weights with UnifiedQA \cite{wortsman2021robust}.


\section{Benchmarking Performance} \label{sec:benchmark}

On a benchmark of 54 real-world binary classification tasks, we show that 1) both re-ranking and fine-tuning are effective, and 2) larger proposers and verifiers are better.

\noindent\textbf{Dataset.} 
The evaluation set of \citet{zhong-etal-2021-adapting-language} aggregated 54 diverse binary text classification tasks, each annotated with one or multiple\footnote{On average 2.2.} natural language descriptions $s^{*}$ for the positive class.
These tasks include topic classification, grammaticality classification, stance classification, etc.
For each task, we asked our systems to describe how the positive class samples differ from the negative class samples and compared the top-5 descriptions the human annotations.

For now, we assume that the annotations $s^*$ are ``correct'' (i.e., the best description to separate the positive and negative classes). 
We will see later that our outputs are sometimes better than $s^*$. 


\noindent\textbf{Evaluated Systems.}
We conjectured that using a larger proposer, a fine-tuned proposer, and a verifier for re-ranking all improve the generated descriptions.
Therefore, we evaluated the following five systems, which all use the verifier from \Cref{sec:fine-tuning} unless otherwise mentioned.
\circled{1}: our hypothetically best system, which uses the fine-tuned GPT-3 Davinci (175B) as the proposer.
\circled{2}: a smaller proposer size (fine-tuned Curie, 13B).
\circled{3}: no fine-tuning (zero-shot Curie, 13B).
\circled{4}: no fine-tuning (zero-shot Curie, 13B), and no verifier for re-ranking.
We also evaluated \circled{5}, a ``memorization proposer'', where the proposer only generates the hypotheses we curated in \Cref{sec:collect};
this ablation makes sure that the fine-tuned proposer's performance is not simply due to memorizing its training set.
If all our conjectures hold, we should find that \circled{1} $>$ \circled{2} $>$ \circled{3} $>$ \circled{4} and \circled{2} $>$ \circled{5}.

\noindent\textbf{Automatic Evaluation.}
We first evaluated our systems using the automatic metric BERTscore \cite{zhang2019bertscore}, which approximates the similarity between two natural language texts.
For each binary task, we computed the BERTscore between every pair of the human annotations and the top-5 descriptions; then, we chose the highest score among all pairs and averaged it across 54 tasks.

Using this metric, we indeed found that \circled{1} (0.930) $>$ \circled{2} (0.927) $>$ \circled{3} (0.907) $>$ \circled{4} (0.899), and \circled{2} (0.927) $>$ \circled{5} (0.916), which validated our conjectures.
However, all these numbers are high, the differences are small, and it is hard to interpret what they imply for the quality of our descriptions.\footnote{\Cref{appendix:use-bert-score} runs a sanity check to make sure that the scores, though not very informative, robustly rank system \circled{1} over \circled{4}.}
Therefore, we additionally evaluated our systems by hand.

\noindent\textbf{Manual Evaluation.}
We evaluated the top-5 descriptions generated for each of the five systems on the 54 binary tasks (total 1350) by hand.
To avoid biases against any of the five systems, the authors were blind towards which system generated each description.
We compared the systems' generated descriptions $\hat{s}$ to human annotations $s^{*}$ and rated their similarity with four levels: 

\begin{table}[]
    \centering
\begin{tabular}{lrrrrr}
\toprule
{} &  \circled{1} best &  \shortstack{\circled{2} smaller} &  \shortstack{\circled{3} no \\ tune} &  \shortstack{\circled{4} no \\verifier} &  \shortstack{\circled{5}\\ memo} \\
\midrule
(A) &                 31 &               22 &     11 &                    4 &                5 \\
(B) &                 10 &               11 &      6 &                    0 &                5 \\
(C) &                  7 &               10 &     10 &                    6 &               21 \\
(D) &                  6 &               11 &     27 &                   44 &               23 \\
\bottomrule
\end{tabular}
\caption{We evaluated each of the five systems as described in \Cref{sec:benchmark}.
\circled{1} largest fine-tuned proposer + verifier, \circled{2} smaller proposer size, \circled{3} no fine-tuining, \circled{4} no re-ranking, and \circled{5} using the memorization proposer.
Better systems have larger numbers in row (A).
Using a larger proposer, a fine-tuned proposer, and a verifier all improve the generated descriptions.
We report the $p$ values in \Cref{appendix:top-k}.
}
\label{tab:main-results}
\end{table}

\begin{itemize}
    \item [(A),] if $\hat{s}$ has mostly the same meaning as one of the human annotations $s^{*}$; e.g., ``\textit{is related to sports}'' = ``\textit{is about sports}.'' 
    \item[(B),] if $\hat{s}$ is close but different; e.g., ``\textit{is about sports team}'' $\approx$ ``\textit{is about sports}.'' 
    \item[(C),] if $\hat{s}$ is highly correlated but has different meaning; for example, 
``\textit{people needs shelter}'' is correlated with ``\textit{there is an earthquake}.'' 
    \item[(D),] if $\hat{s}$ is unrelated to $s^{*}$.
\end{itemize}

For each system, we find the highest rating among the top-5 descriptions and count them across 54 tasks.
We find that for row (A), \circled{1} $>$ \circled{2} $>$ \circled{3} $>$ \circled{4} and \circled{2} $>$ \circled{5}, validating our conjectures.
Adding numbers from row (A) and (B),
we find that while GPT-3 Curie (13B) only generates a description close to human annotation 7\% of the time, the performance reaches 61\% with fine-tuning and re-ranking, and our best system using GPT-3 Davinci (175B) reaches 76\%.
In the appendix, we also present the top-1 performance of our system in \Cref{tab:top-k}, example human annotations, descriptions by our systems, and their ratings in \Cref{tab:example-description}.

Due to resource constraints, we did not systematically investigate whether the verifier is still effective after fine-tuning.
Nevertheless, our qualitative analyses find that the fine-tuned proposer sometimes still generates completely unrelated hypotheses, repeats the hypothesis in the training set, or ``rants'' \footnote{E.g., ``\textit{contains the word ``turned'', which indicates that the weather turned to a certain state}''} based on a specific text sample.
The verifier helps rule them out.
Finally, the proposer has a limited context size and can only generate hypotheses conditioned on five samples, losing information about the entire distribution;
the verifier does not have this fundamental limitation.

\begin{figure}
    \centering
    \includegraphics[width=0.75\columnwidth]{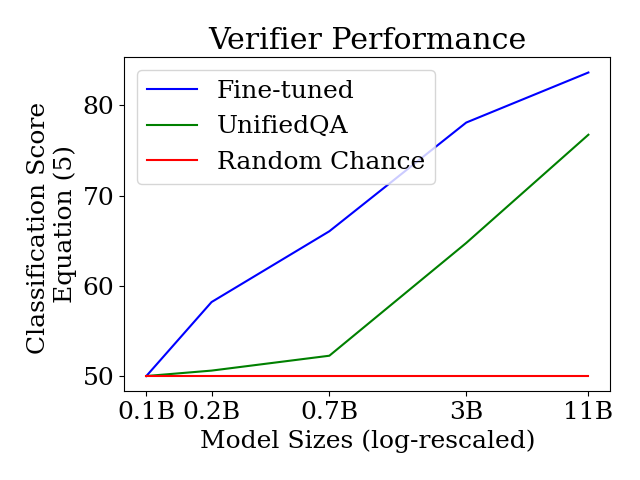}
    \caption{
    We compared verifiers of various sizes and UnifiedQA out of the box by evaluating their binary classification performance, using the metric $\classificationscore(\hat{h}_{s^{*}})$ explained in \Cref{eq:verifier-metrics}.
    We find that fine-tuning and larger model sizes improve the performance.
    }
    \label{fig:verifier_benchmark}
\end{figure}

\noindent\textbf{Comparing Verifiers.}
We next evaluate different choices of the verifier. To test a verifier, we check whether it can reliably separate the two classes when given the gold annotation $h^*$. More precisely, we compute 

\begin{equation} \label{eq:verifier-metrics}
\frac{1}{2} \mathbb{E}_{x_{0}\sim D_{0}, x_{1}\sim D_{1}}[V(s^{*}, x_1, x_0) - V(s^{*}, x_0, x_1) + 1],
\end{equation}

which is equivalent to the classification accuracy $\classificationscore(\hat{h}_{s^{*}})$.

We conjectured that larger and fine-tuned verifiers are better, so we compared our fine-tuned verifier in \Cref{sec:fine-tuning} with smaller ones and UnifiedQA out of the box, averaging $\classificationscore(\hat{h}_{s^{*}})$ across all $54$ tasks.
%
\Cref{fig:verifier_benchmark} visualizes the results.
UnifiedQA performs decently, while additional fine-tuning improves the performance.
Still, $\classificationscore(\hat{h}_{s^{*}})$ is much lower than $1$, implying that re-ranking is imperfect and automatic evaluation by approximating $\classificationscore(h_{s})$ might not yet be feasible.
Nevertheless, these problems may be alleviated in the future: 
the current state of the art models are at least 25x larger than our verifier \cite{ rae2021scaling}, and the curve in \Cref{fig:verifier_benchmark} predicts that their performance will be higher.

\section{Application} \label{sec:application}
We applied our system to summarize training tasks, debug dataset shortcuts, describe distribution shifts, and label text clusters.
All italicized quotes in this section are verbatim generations from our system.

\noindent \textbf{Summarizing Training Tasks.} 
Human descriptions can be imperfect even for widely-used binary classification datasets.
For example, the subjectivity analysis (SUBJ) dataset \cite{pang-lee-2004-sentimental} was proposed as classifying between subjective vs. objective texts, and several works \citep{bragg2021flex, zhong-etal-2021-adapting-language, gao-etal-2021-making, min2021noisy} have used it to test zero/few-shot subjectivity classification.
However, our system generates descriptions ``\textit{is a plot summary of a film}'' for the ``objective'' class and ``\textit{is a quote from a film review}'' for the ``subjective'' class.
We therefore re-read \citet{pang-lee-2004-sentimental} carefully, which says (edited for brevity)
\begin{quote}
    To gather subjective sentences, we collected 5000 movie review snippets from www.rottentomatoes.com. To obtain (mostly) objective data, we took 5000 sentences from plot summaries available from www.imdb.com.
\end{quote}
Therefore, our system's descriptions were in fact more accurate.
We conjecture that similar problems will become increasingly prevalent as the trend of aggregating datasets continues \cite{Mishra2021CrossTaskGV, Sanh2021MultitaskPT}: as datasets come from heterogeneous sources, it is a management challenge to characterize the task of every dataset accurately. Our system may help here.\footnote{
Of course, if our system can already perfectly verify the dataset descriptions by performing the task, then we might not need those datasets for training in the first place.
However, even an imperfect AI system can help correct some human mistakes.
}

\noindent \textbf{Debugging Dataset Shortcuts.}
Datasets frequently contain unintended shortcuts. 
For example, the task of natural language inference (NLI) is to verify whether a \texttt{hypothesis}\footnote{This is an NLI-specific concept; we use a special font to distinguish it from ``hypothesis'' (\Cref{sec:definition}) in our paper.} is an entailment or a contradiction given a premise.
The popular MNLI \cite{williams-etal-2018-broad} dataset contains a spurious correlation between contradictions and negations (``not'', ``never'', etc.), and some models learn to predict a contradiction whenever these expressions occur, regardless of the premise \cite{gururangan-etal-2018-annotation}.

If we know what shortcuts are present, we can apply fixes like group DRO \cite{Sagawa2019DistributionallyRN}. But how do we find them in the first place?
We used our system to look for (alternative) descriptions of the differences between the two classes.
We fed the \texttt{hypotheses} from the entailment class and those from the contradiction class to our system, which responded with ``\textit{contains a negative statement}''/``\textit{has a negative verb},'' revealing the spurious shortcut.

We also applied our system to a popular spam classification dataset \cite{gomez2006content}.
We fed sentences from the two classes to our system, which tells us that the spam group ``\textit{has a higher number of hyperlinks}.''
To test whether such URLs influence downstream classifiers, we fed ten of our research communication messages with URLs
to a RoBERTa-Large \cite{Liu2019RoBERTaAR} model fine-tuned on this dataset (99\% in-distribution accuracy).
All 10 messages with URLs were classified as spam and were all classified as non-spam after removing the URLs.

\noindent \textbf{Describing Distribution Shifts.}
We applied our system to describe distribution shifts for natural language tasks. 
For example, in contrast to MNLI, the SNLI dataset \cite{bowman-etal-2015-large} is based on image captions; therefore, our system says that SNLI ``\textit{describes a picture}.''
\citet{naik-etal-2018-stress} constructed another NLI dataset to stress test models' numerical reasoning ability; therefore, our system says that it ``\textit{contains a higher number of number words}.''
To take a different task, TwitterPPDB \cite{lan-etal-2017-continuously} and QQP\footnote{https://www.kaggle.com/c/quora-question-pairs} are both paraphrase detection datasets; the former is constructed by tweets while the latter is constructed by Quora questions;
therefore, the system says that the former ``\textit{talks about a news story more}'' while the latter ``\textit{contains a question}.''

\begin{figure}
    \centering
    \includegraphics[width=0.75\columnwidth]{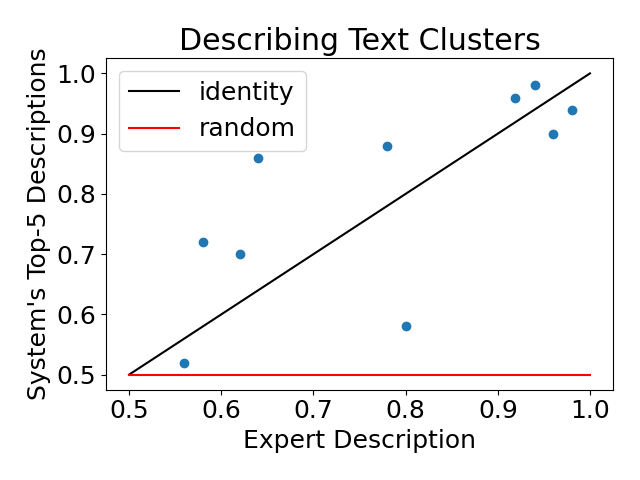}
    \caption{
    For each text cluster (dot), we collect human annotations to compute $\classificationscore(h_{s})$ for the descriptions by our expert ($x$-axis) and the top-5 by our system ($y$-axis). Our system is on par with the expert most of the time.
    }
    \label{fig:text-clusters}
\end{figure}

\noindent \textbf{Labelling Text Clusters.} 
Unsupervised algorithms generate semantically meaningful text clusters; however, researchers usually need to manually examine each of them to identify its semantics \cite{chang2009reading}.
Our system can automatically describe a text cluster by treating it as $D_{1}$ and all others as $D_{0}$.

We compared our system to an expert on their ability to describe clusters.
To create the clusters, we used RoBERTa-Base to embed the test set of wikitext-2 \citep{merity2016pointer} (9992 sentences) and use the approach of \citet{aharoni2020unsupervised} to create 64 clusters.
We randomly selected ten of them for evaluation;
for each of them, one of our authors read through 20 samples and wrote a natural language description $s^{*}$;
we then asked him to read the top-5 descriptions by our system and pick the one $\hat{s}$ that he considered to be the best.
We evaluated this author's performance by $\classificationscore (h_{s^{*}})$ and our system's performance by $\classificationscore (h_{\hat{s}})$, where we collected MTurks' annotations to compute $h_{s}(x_{0}, x_{1})$.

Averaged across all clusters, our system achieves $\classificationscore$=0.8 while the expert achieves 0.77.
Figure \ref{fig:text-clusters} shows the results for each cluster, and we found that our system is at least on par with the expert most of the time.

\textbf{Discussion.} In all the above applications, our system only informs the decisions of the stakeholders, who have the ultimate responsibility to decide if subjectivity can be approximated by ``being review like'', if specific correlations are bugs, or if the distribution shift is severe enough to take action.
Our system also needs to improve to handle these applications robustly. 
For example, in the SPAM classification application, our verifier cannot verify whether a hyperlink exists as reliably as a rule-based classifier, while the 16x larger proposer does the heavy lifting.
We hope scaling up can alleviate this problem in the future.

\section{Related Work}

\noindent\textbf{Prompting and Zero-Shot Learning.}
Checking whether a hypothesis holds for a piece of text can be formulated as a Natural Language Inference \cite{bowman-etal-2015-large} or a Question Answering \cite{clark-etal-2019-boolq} task. 
Recent large pre-trained models can generalize to hypotheses significantly outside the training set \cite{khashabi-etal-2020-unifiedqa}, which allows us to re-rank candidate hypotheses.
We expect future verifiers to be stronger as model sizes and the number of fine-tuned tasks grow \cite{Wei2021FinetunedLM, Sanh2021MultitaskPT}.

Our paper does not search for prompts to improve target task accuracy as in  \citet{shin-etal-2020-autoprompt, mishra2021reframing, rubin2021learning}, which typically assume the target task is known or do not enforce prompt interpretability.
Nevertheless, cross-pollination of ideas might be helpful.

To propose hypotheses, another plausible strategy is to find a continuous prompt \cite{li-liang-2021-prefix} first and then decode it to natural language by adding perplexity constraints \cite{song-etal-2020-adversarial}.
However, \citet{khashabi2021prompt} suggests that this might be hard, given that soft prompts are not unique and heavily depend on initialization.

\noindent\textbf{AI Safety and Scalable Oversight.} Machine learning algorithms often fail on input patterns that are rare during train time.
Typical examples include out-of-distribution samples \cite{hendrycks2021many}, unforeseen adversaries \cite{kang2019testing}, spurious shortcuts \cite{sagawa2019distributionally}, and their interactions with the target population \cite{hardt2016strategic, hashimoto2018fairness}. 
Our system can monitor the differences between the train and test distribution to inform decision-makers. 
More broadly, we hope our automatically generated descriptions can help humans scalably oversee complicated machine learning systems \cite{amodei2016concrete}. 

\noindent\textbf{Learning a Predictor as Explanation.}
It is not new to discover statistical relationships in data by interpreting a learned hypothesis.
Given real-valued features and a target variable, economists frequently run linear regressions and analyze the effect of each feature by interpreting the learned weights \cite{draper1998applied}, sometimes adding sparsity constraints to focus on more important ones \citep{342465, abbasiasl2020structural}.
Decision tree with a small list of if-then statements can also extract interpretable rules, e.g., to predict strokes \citep{2015}.
In comparison, our work focuses on discovering patterns in structured data (e.g., text) rather than real vectors; we also learn a natural language description, which might be easier for humans to understand, rather than a mathematical expression.

\section{Discussion} \label{sec:disussion}

\textbf{Directions for Improving.} 
Besides increasing model sizes (\Cref{sec:benchmark}), our method would also benefit from: 1) running the proposer on different sets of samples and ensembling their outputs \cite{min2021noisy},
2) using a proposer with a larger context size \cite{Kitaev2020ReformerTE}, 3) using a verifier with a symbolic component for numerical computations \cite{cobbe2021training}, and 4) using a retriever to verify information from external sources \cite{nakano2021webgpt}.  
Additionally, \Cref{appendix:method-remark} interprets our method under a unifying probabilistic framework and discusses future directions using cycle consistency and self-supervision.

We currently evaluate only 54 distribution pairs by hand, which is time-consuming and small in scale.
This might prevent future researchers from validating new methods quickly and reliably.
We hope that automatic metrics more discriminative than \citet{zhang2019bertscore} will help in the future, and that the number of distribution pairs for evaluation will increase as the community continues pooling datasets together \cite{Mishra2021CrossTaskGV, Sanh2021MultitaskPT}. 

\textbf{Inherent Ambiguities in Natural Language.}
Classical statistical analyses usually study mathematical hypotheses, whose meaning is uncontroversial and never changes; for example, people from different cultures and eras would all agree on what the number ``7'' means.
However, there is no canonical way to interpret a natural language hypothesis:
for example, \citet{sap2021annotators} finds that annotators with different social backgrounds disagree on the meaning of ``\textit{this sentence is offensive}.''
Future systems need to consider the listeners' background to prevent biases and ambiguities.

\textbf{Expressiveness of the Descriptions.}
Our work only considers descriptions in the form of short natural language sentences. 
However, a single short sentence is sometimes insufficient to capture the multifaceted differences between two distributions, and hence multiple different descriptions are plausible. 
Currently, the user can choose to examine an arbitrary number of our systems' proposed hypotheses, sorted by \Cref{eq:verifier-metrics}. 
Future work may consider logical compositions (e.g., conjunctions) of multiple hypotheses, hence making the descriptions more expressive.

Theoretically, natural language descriptions could be more expressive than those demonstrated in this paper, given that most of the human knowledge is communicated through natural language. 
However, they can still be limited, since humans can know more than they can verbalize (Polanyi's paradox, \cite{polanyi2009tacit}).

\textbf{Broader Applications.}
Our paper only considers text distributions, but language can also describe other modalities, such as vision \cite{radford2021learning}, sound \cite{barchiesi2015acoustic},  smell \cite{kiela2015grounding}, taste \cite{nozaki2018predictive}, or motor sensations \cite{thomason2016learning}.
In principle, our framework can adapt to any experience humans can describe through language.

Our framework can also help answer broader scientific questions, for example: 
what does individual neuron represent in a deep learning model \cite{hernandez2022natural}, 
how people from different parties discuss shooting events \cite{Demszky2019AnalyzingPI}, how people with different psychological signatures write \cite{boyd2015did}, or how search queries change over time \cite{gentzkow2019text}.\footnote{\citet{zeng2002modeling} note that the volume of searches or web hits seeking information related to a disease may be a strong predictor of its prevalence.}
We hope our method can help humans scalably discover new patterns in big data and complex systems.

\section*{Acknowledgement}
The first author is funded by NSF-Simons Theorinet Grant (NSF Award \#2031985). 
We thank OpenAI for providing inference and fine-tuning access to GPT-3 Davinci.
We thank the Berkeley NLP group, the Steinhardt Group, and the anonymous reviewers for their feedbacks on the paper. 
We thank Dong Yang for training a new proposer by fine-tuning T5, and the TPU Research Cloud (TRC) program for providing computational resources. 

\nocite{langley00}

\bibliography{example_paper}

\begin{thebibliography}{67}
\providecommand{\natexlab}[1]{#1}
\providecommand{\url}[1]{\texttt{#1}}
\expandafter\ifx\csname urlstyle\endcsname\relax
  \providecommand{\doi}[1]{doi: #1}\else
  \providecommand{\doi}{doi: \begingroup \urlstyle{rm}\Url}\fi

\bibitem[Abbasi-Asl \& Yu(2020)Abbasi-Asl and Yu]{abbasiasl2020structural}
Abbasi-Asl, R. and Yu, B.
\newblock Structural compression of convolutional neural networks, 2020.

\bibitem[Aharoni \& Goldberg(2020)Aharoni and
  Goldberg]{aharoni2020unsupervised}
Aharoni, R. and Goldberg, Y.
\newblock Unsupervised domain clusters in pretrained language models, 2020.

\bibitem[Almeida et~al.(2013)Almeida, Hidalgo, and Silva]{almeida2013towards}
Almeida, T., Hidalgo, J. M.~G., and Silva, T.~P.
\newblock Towards sms spam filtering: Results under a new dataset.
\newblock \emph{International Journal of Information Security Science},
  2\penalty0 (1):\penalty0 1--18, 2013.

\bibitem[Amodei et~al.(2016)Amodei, Olah, Steinhardt, Christiano, Schulman, and
  Man{\'e}]{amodei2016concrete}
Amodei, D., Olah, C., Steinhardt, J., Christiano, P., Schulman, J., and
  Man{\'e}, D.
\newblock Concrete problems in ai safety.
\newblock \emph{arXiv preprint arXiv:1606.06565}, 2016.

\bibitem[Barbieri et~al.(2020)Barbieri, Camacho-Collados, Espinosa~Anke, and
  Neves]{barbieri-etal-2020-tweeteval}
Barbieri, F., Camacho-Collados, J., Espinosa~Anke, L., and Neves, L.
\newblock {T}weet{E}val: Unified benchmark and comparative evaluation for tweet
  classification.
\newblock In \emph{Findings of the Association for Computational Linguistics:
  EMNLP 2020}, pp.\  1644--1650, Online, November 2020. Association for
  Computational Linguistics.
\newblock \doi{10.18653/v1/2020.findings-emnlp.148}.
\newblock URL \url{https://aclanthology.org/2020.findings-emnlp.148}.

\bibitem[Barchiesi et~al.(2015)Barchiesi, Giannoulis, Stowell, and
  Plumbley]{barchiesi2015acoustic}
Barchiesi, D., Giannoulis, D., Stowell, D., and Plumbley, M.~D.
\newblock Acoustic scene classification: Classifying environments from the
  sounds they produce.
\newblock \emph{IEEE Signal Processing Magazine}, 32\penalty0 (3):\penalty0
  16--34, 2015.

\bibitem[Bowman et~al.(2015)Bowman, Angeli, Potts, and
  Manning]{bowman-etal-2015-large}
Bowman, S.~R., Angeli, G., Potts, C., and Manning, C.~D.
\newblock A large annotated corpus for learning natural language inference.
\newblock In \emph{Proceedings of the 2015 Conference on Empirical Methods in
  Natural Language Processing}, pp.\  632--642, Lisbon, Portugal, September
  2015. Association for Computational Linguistics.
\newblock \doi{10.18653/v1/D15-1075}.
\newblock URL \url{https://aclanthology.org/D15-1075}.

\bibitem[Boyd \& Pennebaker(2015)Boyd and Pennebaker]{boyd2015did}
Boyd, R.~L. and Pennebaker, J.~W.
\newblock Did shakespeare write double falsehood? identifying individuals by
  creating psychological signatures with text analysis.
\newblock \emph{Psychological science}, 26\penalty0 (5):\penalty0 570--582,
  2015.

\bibitem[Bragg et~al.(2021)Bragg, Cohan, Lo, and Beltagy]{bragg2021flex}
Bragg, J., Cohan, A., Lo, K., and Beltagy, I.
\newblock Flex: Unifying evaluation for few-shot nlp.
\newblock \emph{Advances in Neural Information Processing Systems}, 34, 2021.

\bibitem[Brown et~al.(2020)Brown, Mann, Ryder, Subbiah, Kaplan, Dhariwal,
  Neelakantan, Shyam, Sastry, Askell, et~al.]{brown2020language}
Brown, T.~B., Mann, B., Ryder, N., Subbiah, M., Kaplan, J., Dhariwal, P.,
  Neelakantan, A., Shyam, P., Sastry, G., Askell, A., et~al.
\newblock Language models are few-shot learners.
\newblock \emph{arXiv preprint arXiv:2005.14165}, 2020.

\bibitem[Chang et~al.(2009)Chang, Gerrish, Wang, Boyd-Graber, and
  Blei]{chang2009reading}
Chang, J., Gerrish, S., Wang, C., Boyd-Graber, J., and Blei, D.
\newblock Reading tea leaves: How humans interpret topic models.
\newblock \emph{Advances in neural information processing systems}, 22, 2009.

\bibitem[Clark et~al.(2019)Clark, Lee, Chang, Kwiatkowski, Collins, and
  Toutanova]{clark-etal-2019-boolq}
Clark, C., Lee, K., Chang, M.-W., Kwiatkowski, T., Collins, M., and Toutanova,
  K.
\newblock {B}ool{Q}: Exploring the surprising difficulty of natural yes/no
  questions.
\newblock In \emph{Proceedings of the 2019 Conference of the North {A}merican
  Chapter of the Association for Computational Linguistics: Human Language
  Technologies, Volume 1 (Long and Short Papers)}, pp.\  2924--2936,
  Minneapolis, Minnesota, June 2019. Association for Computational Linguistics.
\newblock \doi{10.18653/v1/N19-1300}.
\newblock URL \url{https://aclanthology.org/N19-1300}.

\bibitem[Cobbe et~al.(2021)Cobbe, Kosaraju, Bavarian, Hilton, Nakano, Hesse,
  and Schulman]{cobbe2021training}
Cobbe, K., Kosaraju, V., Bavarian, M., Hilton, J., Nakano, R., Hesse, C., and
  Schulman, J.
\newblock Training verifiers to solve math word problems.
\newblock \emph{arXiv preprint arXiv:2110.14168}, 2021.

\bibitem[Demszky et~al.(2019)Demszky, Garg, Voigt, Zou, Gentzkow, Shapiro, and
  Jurafsky]{Demszky2019AnalyzingPI}
Demszky, D., Garg, N., Voigt, R., Zou, J.~Y., Gentzkow, M., Shapiro, J.~M., and
  Jurafsky, D.
\newblock Analyzing polarization in social media: Method and application to
  tweets on 21 mass shootings.
\newblock In \emph{NAACL}, 2019.

\bibitem[Draper \& Smith(1998)Draper and Smith]{draper1998applied}
Draper, N.~R. and Smith, H.
\newblock \emph{Applied regression analysis}, volume 326.
\newblock John Wiley \& Sons, 1998.

\bibitem[Gao et~al.(2021)Gao, Fisch, and Chen]{gao-etal-2021-making}
Gao, T., Fisch, A., and Chen, D.
\newblock Making pre-trained language models better few-shot learners.
\newblock In \emph{Proceedings of the 59th Annual Meeting of the Association
  for Computational Linguistics and the 11th International Joint Conference on
  Natural Language Processing (Volume 1: Long Papers)}, pp.\  3816--3830,
  Online, August 2021. Association for Computational Linguistics.
\newblock \doi{10.18653/v1/2021.acl-long.295}.
\newblock URL \url{https://aclanthology.org/2021.acl-long.295}.

\bibitem[Gentzkow et~al.(2019)Gentzkow, Kelly, and Taddy]{gentzkow2019text}
Gentzkow, M., Kelly, B., and Taddy, M.
\newblock Text as data.
\newblock \emph{Journal of Economic Literature}, 57\penalty0 (3):\penalty0
  535--74, 2019.

\bibitem[G{\'o}mez~Hidalgo et~al.(2006)G{\'o}mez~Hidalgo, Bringas, S{\'a}nz,
  and Garc{\'\i}a]{gomez2006content}
G{\'o}mez~Hidalgo, J.~M., Bringas, G.~C., S{\'a}nz, E.~P., and Garc{\'\i}a,
  F.~C.
\newblock Content based sms spam filtering.
\newblock In \emph{Proceedings of the 2006 ACM symposium on Document
  engineering}, pp.\  107--114, 2006.

\bibitem[Gururangan et~al.(2018)Gururangan, Swayamdipta, Levy, Schwartz,
  Bowman, and Smith]{gururangan-etal-2018-annotation}
Gururangan, S., Swayamdipta, S., Levy, O., Schwartz, R., Bowman, S., and Smith,
  N.~A.
\newblock Annotation artifacts in natural language inference data.
\newblock In \emph{Proceedings of the 2018 Conference of the North {A}merican
  Chapter of the Association for Computational Linguistics: Human Language
  Technologies, Volume 2 (Short Papers)}, pp.\  107--112, New Orleans,
  Louisiana, June 2018. Association for Computational Linguistics.
\newblock \doi{10.18653/v1/N18-2017}.
\newblock URL \url{https://aclanthology.org/N18-2017}.

\bibitem[Hardt et~al.(2016)Hardt, Megiddo, Papadimitriou, and
  Wootters]{hardt2016strategic}
Hardt, M., Megiddo, N., Papadimitriou, C., and Wootters, M.
\newblock Strategic classification.
\newblock In \emph{Proceedings of the 2016 ACM conference on innovations in
  theoretical computer science}, pp.\  111--122, 2016.

\bibitem[Hashimoto et~al.(2018)Hashimoto, Srivastava, Namkoong, and
  Liang]{hashimoto2018fairness}
Hashimoto, T., Srivastava, M., Namkoong, H., and Liang, P.
\newblock Fairness without demographics in repeated loss minimization.
\newblock In \emph{International Conference on Machine Learning}, pp.\
  1929--1938. PMLR, 2018.

\bibitem[Hendrycks et~al.(2021)Hendrycks, Basart, Mu, Kadavath, Wang, Dorundo,
  Desai, Zhu, Parajuli, Guo, et~al.]{hendrycks2021many}
Hendrycks, D., Basart, S., Mu, N., Kadavath, S., Wang, F., Dorundo, E., Desai,
  R., Zhu, T., Parajuli, S., Guo, M., et~al.
\newblock The many faces of robustness: A critical analysis of
  out-of-distribution generalization.
\newblock In \emph{Proceedings of the IEEE/CVF International Conference on
  Computer Vision}, pp.\  8340--8349, 2021.

\bibitem[Hernandez et~al.(2022)Hernandez, Schwettmann, Bau, Bagashvili,
  Torralba, and Andreas]{hernandez2022natural}
Hernandez, E., Schwettmann, S., Bau, D., Bagashvili, T., Torralba, A., and
  Andreas, J.
\newblock Natural language descriptions of deep visual features.
\newblock In \emph{International Conference on Learning Representations}, 2022.
\newblock URL \url{https://arxiv.org/abs/2201.11114}.

\bibitem[Kang et~al.(2019)Kang, Sun, Hendrycks, Brown, and
  Steinhardt]{kang2019testing}
Kang, D., Sun, Y., Hendrycks, D., Brown, T., and Steinhardt, J.
\newblock Testing robustness against unforeseen adversaries.
\newblock \emph{arXiv preprint arXiv:1908.08016}, 2019.

\bibitem[Khashabi et~al.(2020)Khashabi, Min, Khot, Sabharwal, Tafjord, Clark,
  and Hajishirzi]{khashabi-etal-2020-unifiedqa}
Khashabi, D., Min, S., Khot, T., Sabharwal, A., Tafjord, O., Clark, P., and
  Hajishirzi, H.
\newblock {UNIFIEDQA}: Crossing format boundaries with a single {QA} system.
\newblock In \emph{Findings of the Association for Computational Linguistics:
  EMNLP 2020}, pp.\  1896--1907, Online, November 2020. Association for
  Computational Linguistics.
\newblock \doi{10.18653/v1/2020.findings-emnlp.171}.
\newblock URL \url{https://aclanthology.org/2020.findings-emnlp.171}.

\bibitem[Khashabi et~al.(2021)Khashabi, Lyu, Min, Qin, Richardson, Singh,
  Welleck, Hajishirzi, Khot, Sabharwal, et~al.]{khashabi2021prompt}
Khashabi, D., Lyu, S., Min, S., Qin, L., Richardson, K., Singh, S., Welleck,
  S., Hajishirzi, H., Khot, T., Sabharwal, A., et~al.
\newblock Prompt waywardness: The curious case of discretized interpretation of
  continuous prompts.
\newblock \emph{arXiv preprint arXiv:2112.08348}, 2021.

\bibitem[Kiela et~al.(2015)Kiela, Bulat, and Clark]{kiela2015grounding}
Kiela, D., Bulat, L., and Clark, S.
\newblock Grounding semantics in olfactory perception.
\newblock In \emph{Proceedings of the 53rd Annual Meeting of the Association
  for Computational Linguistics and the 7th International Joint Conference on
  Natural Language Processing (Volume 2: Short Papers)}, pp.\  231--236, 2015.

\bibitem[Kitaev et~al.(2020)Kitaev, Kaiser, and Levskaya]{Kitaev2020ReformerTE}
Kitaev, N., Kaiser, L., and Levskaya, A.
\newblock Reformer: The efficient transformer.
\newblock \emph{ArXiv}, abs/2001.04451, 2020.

\bibitem[Lan et~al.(2017)Lan, Qiu, He, and Xu]{lan-etal-2017-continuously}
Lan, W., Qiu, S., He, H., and Xu, W.
\newblock A continuously growing dataset of sentential paraphrases.
\newblock In \emph{Proceedings of the 2017 Conference on Empirical Methods in
  Natural Language Processing}, pp.\  1224--1234, Copenhagen, Denmark,
  September 2017. Association for Computational Linguistics.
\newblock \doi{10.18653/v1/D17-1126}.
\newblock URL \url{https://aclanthology.org/D17-1126}.

\bibitem[Langley(2000)]{langley00}
Langley, P.
\newblock Crafting papers on machine learning.
\newblock In Langley, P. (ed.), \emph{Proceedings of the 17th International
  Conference on Machine Learning (ICML 2000)}, pp.\  1207--1216, Stanford, CA,
  2000. Morgan Kaufmann.

\bibitem[Letham et~al.(2015)Letham, Rudin, McCormick, and Madigan]{2015}
Letham, B., Rudin, C., McCormick, T.~H., and Madigan, D.
\newblock Interpretable classifiers using rules and bayesian analysis: Building
  a better stroke prediction model.
\newblock \emph{The Annals of Applied Statistics}, 9\penalty0 (3), Sep 2015.
\newblock ISSN 1932-6157.
\newblock \doi{10.1214/15-aoas848}.
\newblock URL \url{http://dx.doi.org/10.1214/15-AOAS848}.

\bibitem[Li \& Roth(2002)Li and Roth]{li2002learning}
Li, X. and Roth, D.
\newblock Learning question classifiers.
\newblock In \emph{COLING 2002: The 19th International Conference on
  Computational Linguistics}, 2002.

\bibitem[Li \& Liang(2021)Li and Liang]{li-liang-2021-prefix}
Li, X.~L. and Liang, P.
\newblock Prefix-tuning: Optimizing continuous prompts for generation.
\newblock In \emph{Proceedings of the 59th Annual Meeting of the Association
  for Computational Linguistics and the 11th International Joint Conference on
  Natural Language Processing (Volume 1: Long Papers)}, pp.\  4582--4597,
  Online, August 2021. Association for Computational Linguistics.
\newblock \doi{10.18653/v1/2021.acl-long.353}.
\newblock URL \url{https://aclanthology.org/2021.acl-long.353}.

\bibitem[Liu et~al.(2019)Liu, Ott, Goyal, Du, Joshi, Chen, Levy, Lewis,
  Zettlemoyer, and Stoyanov]{Liu2019RoBERTaAR}
Liu, Y., Ott, M., Goyal, N., Du, J., Joshi, M., Chen, D., Levy, O., Lewis, M.,
  Zettlemoyer, L., and Stoyanov, V.
\newblock Roberta: A robustly optimized bert pretraining approach.
\newblock \emph{ArXiv}, abs/1907.11692, 2019.

\bibitem[Maas et~al.(2011)Maas, Daly, Pham, Huang, Ng, and
  Potts]{maas-EtAl:2011:ACL-HLT2011}
Maas, A.~L., Daly, R.~E., Pham, P.~T., Huang, D., Ng, A.~Y., and Potts, C.
\newblock Learning word vectors for sentiment analysis.
\newblock In \emph{Proceedings of the 49th Annual Meeting of the Association
  for Computational Linguistics: Human Language Technologies}, pp.\  142--150,
  Portland, Oregon, USA, June 2011. Association for Computational Linguistics.
\newblock URL \url{http://www.aclweb.org/anthology/P11-1015}.

\bibitem[Merity et~al.(2016)Merity, Xiong, Bradbury, and
  Socher]{merity2016pointer}
Merity, S., Xiong, C., Bradbury, J., and Socher, R.
\newblock Pointer sentinel mixture models, 2016.

\bibitem[Mihaylova et~al.(2019)Mihaylova, Karadzhov, Atanasova, Baly,
  Mohtarami, and Nakov]{mihaylova-etal-2019-semeval}
Mihaylova, T., Karadzhov, G., Atanasova, P., Baly, R., Mohtarami, M., and
  Nakov, P.
\newblock {S}em{E}val-2019 task 8: Fact checking in community question
  answering forums.
\newblock In \emph{Proceedings of the 13th International Workshop on Semantic
  Evaluation}, pp.\  860--869, Minneapolis, Minnesota, USA, June 2019.
  Association for Computational Linguistics.
\newblock \doi{10.18653/v1/S19-2149}.
\newblock URL \url{https://aclanthology.org/S19-2149}.

\bibitem[Min et~al.(2021)Min, Lewis, Hajishirzi, and Zettlemoyer]{min2021noisy}
Min, S., Lewis, M., Hajishirzi, H., and Zettlemoyer, L.
\newblock Noisy channel language model prompting for few-shot text
  classification.
\newblock \emph{arXiv preprint arXiv:2108.04106}, 2021.

\bibitem[Mishra et~al.(2021{\natexlab{a}})Mishra, Khashabi, Baral, Choi, and
  Hajishirzi]{mishra2021reframing}
Mishra, S., Khashabi, D., Baral, C., Choi, Y., and Hajishirzi, H.
\newblock Reframing instructional prompts to gptk's language.
\newblock \emph{arXiv preprint arXiv:2109.07830}, 2021{\natexlab{a}}.

\bibitem[Mishra et~al.(2021{\natexlab{b}})Mishra, Khashabi, Baral, and
  Hajishirzi]{Mishra2021CrossTaskGV}
Mishra, S., Khashabi, D., Baral, C., and Hajishirzi, H.
\newblock Cross-task generalization via natural language crowdsourcing
  instructions.
\newblock 2021{\natexlab{b}}.

\bibitem[Naik et~al.(2018)Naik, Ravichander, Sadeh, Rose, and
  Neubig]{naik-etal-2018-stress}
Naik, A., Ravichander, A., Sadeh, N., Rose, C., and Neubig, G.
\newblock Stress test evaluation for natural language inference.
\newblock In \emph{Proceedings of the 27th International Conference on
  Computational Linguistics}, pp.\  2340--2353, Santa Fe, New Mexico, USA,
  August 2018. Association for Computational Linguistics.
\newblock URL \url{https://aclanthology.org/C18-1198}.

\bibitem[Nakano et~al.(2021)Nakano, Hilton, Balaji, Wu, Ouyang, Kim, Hesse,
  Jain, Kosaraju, Saunders, et~al.]{nakano2021webgpt}
Nakano, R., Hilton, J., Balaji, S., Wu, J., Ouyang, L., Kim, C., Hesse, C.,
  Jain, S., Kosaraju, V., Saunders, W., et~al.
\newblock Webgpt: Browser-assisted question-answering with human feedback.
\newblock \emph{arXiv preprint arXiv:2112.09332}, 2021.

\bibitem[Nozaki \& Nakamoto(2018)Nozaki and Nakamoto]{nozaki2018predictive}
Nozaki, Y. and Nakamoto, T.
\newblock Predictive modeling for odor character of a chemical using machine
  learning combined with natural language processing.
\newblock \emph{PloS one}, 13\penalty0 (6):\penalty0 e0198475, 2018.

\bibitem[Pang \& Lee(2004)Pang and Lee]{pang-lee-2004-sentimental}
Pang, B. and Lee, L.
\newblock A sentimental education: Sentiment analysis using subjectivity
  summarization based on minimum cuts.
\newblock In \emph{Proceedings of the 42nd Annual Meeting of the Association
  for Computational Linguistics ({ACL}-04)}, pp.\  271--278, Barcelona, Spain,
  July 2004.
\newblock \doi{10.3115/1218955.1218990}.
\newblock URL \url{https://aclanthology.org/P04-1035}.

\bibitem[Pati et~al.(1993)Pati, Rezaiifar, and Krishnaprasad]{342465}
Pati, Y., Rezaiifar, R., and Krishnaprasad, P.
\newblock Orthogonal matching pursuit: recursive function approximation with
  applications to wavelet decomposition.
\newblock In \emph{Proceedings of 27th Asilomar Conference on Signals, Systems
  and Computers}, pp.\  40--44 vol.1, 1993.
\newblock \doi{10.1109/ACSSC.1993.342465}.

\bibitem[Polanyi \& Sen(2009)Polanyi and Sen]{polanyi2009tacit}
Polanyi, M. and Sen, A.
\newblock \emph{The tacit dimension}.
\newblock University of Chicago press, 2009.

\bibitem[Radford et~al.(2021)Radford, Kim, Hallacy, Ramesh, Goh, Agarwal,
  Sastry, Askell, Mishkin, Clark, et~al.]{radford2021learning}
Radford, A., Kim, J.~W., Hallacy, C., Ramesh, A., Goh, G., Agarwal, S., Sastry,
  G., Askell, A., Mishkin, P., Clark, J., et~al.
\newblock Learning transferable visual models from natural language
  supervision.
\newblock \emph{arXiv preprint arXiv:2103.00020}, 2021.

\bibitem[Rae et~al.(2021)Rae, Borgeaud, Cai, Millican, Hoffmann, Song,
  Aslanides, Henderson, Ring, Young, et~al.]{rae2021scaling}
Rae, J.~W., Borgeaud, S., Cai, T., Millican, K., Hoffmann, J., Song, F.,
  Aslanides, J., Henderson, S., Ring, R., Young, S., et~al.
\newblock Scaling language models: Methods, analysis \& insights from training
  gopher.
\newblock \emph{arXiv preprint arXiv:2112.11446}, 2021.

\bibitem[Raffel et~al.(2019)Raffel, Shazeer, Roberts, Lee, Narang, Matena,
  Zhou, Li, and Liu]{raffel2019exploring}
Raffel, C., Shazeer, N., Roberts, A., Lee, K., Narang, S., Matena, M., Zhou,
  Y., Li, W., and Liu, P.~J.
\newblock Exploring the limits of transfer learning with a unified text-to-text
  transformer.
\newblock \emph{arXiv preprint arXiv:1910.10683}, 2019.

\bibitem[Rubin et~al.(2021)Rubin, Herzig, and Berant]{rubin2021learning}
Rubin, O., Herzig, J., and Berant, J.
\newblock Learning to retrieve prompts for in-context learning.
\newblock \emph{arXiv preprint arXiv:2112.08633}, 2021.

\bibitem[Sagawa et~al.(2019{\natexlab{a}})Sagawa, Koh, Hashimoto, and
  Liang]{Sagawa2019DistributionallyRN}
Sagawa, S., Koh, P.~W., Hashimoto, T.~B., and Liang, P.
\newblock Distributionally robust neural networks for group shifts: On the
  importance of regularization for worst-case generalization.
\newblock \emph{ArXiv}, abs/1911.08731, 2019{\natexlab{a}}.

\bibitem[Sagawa et~al.(2019{\natexlab{b}})Sagawa, Koh, Hashimoto, and
  Liang]{sagawa2019distributionally}
Sagawa, S., Koh, P.~W., Hashimoto, T.~B., and Liang, P.
\newblock Distributionally robust neural networks for group shifts: On the
  importance of regularization for worst-case generalization.
\newblock \emph{arXiv preprint arXiv:1911.08731}, 2019{\natexlab{b}}.

\bibitem[Sanh et~al.(2021)Sanh, Webson, Raffel, Bach, Sutawika, Alyafeai,
  Chaffin, Stiegler, Scao, Raja, Dey, BARI, Xu, Thakker, Sharma, Szczechla,
  Kim, Chhablani, Nayak, Datta, Chang, Jiang, Wang, Manica, Shen, Yong, Pandey,
  Bawden, Wang, Neeraj, Rozen, Sharma, Santilli, F{\'e}vry, Fries, Teehan,
  Biderman, Gao, Bers, Wolf, and Rush]{Sanh2021MultitaskPT}
Sanh, V., Webson, A., Raffel, C., Bach, S.~H., Sutawika, L.~A., Alyafeai, Z.,
  Chaffin, A., Stiegler, A., Scao, T.~L., Raja, A., Dey, M., BARI, M.~S., Xu,
  C., Thakker, U., Sharma, S.~S., Szczechla, E., Kim, T., Chhablani, G., Nayak,
  N.~V., Datta, D., Chang, J., Jiang, M. T.-J., Wang, H., Manica, M., Shen, S.,
  Yong, Z.~X., Pandey, H., Bawden, R., Wang, T., Neeraj, T., Rozen, J., Sharma,
  A., Santilli, A., F{\'e}vry, T., Fries, J.~A., Teehan, R., Biderman, S.~R.,
  Gao, L., Bers, T. G.~O., Wolf, T., and Rush, A.~M.
\newblock Multitask prompted training enables zero-shot task generalization.
\newblock \emph{ArXiv}, abs/2110.08207, 2021.

\bibitem[Sap et~al.(2021)Sap, Swayamdipta, Vianna, Zhou, Choi, and
  Smith]{sap2021annotators}
Sap, M., Swayamdipta, S., Vianna, L., Zhou, X., Choi, Y., and Smith, N.~A.
\newblock Annotators with attitudes: How annotator beliefs and identities bias
  toxic language detection.
\newblock \emph{arXiv preprint arXiv:2111.07997}, 2021.

\bibitem[Shin et~al.(2020)Shin, Razeghi, Logan~IV, Wallace, and
  Singh]{shin-etal-2020-autoprompt}
Shin, T., Razeghi, Y., Logan~IV, R.~L., Wallace, E., and Singh, S.
\newblock {A}uto{P}rompt: {E}liciting {K}nowledge from {L}anguage {M}odels with
  {A}utomatically {G}enerated {P}rompts.
\newblock In \emph{Proceedings of the 2020 Conference on Empirical Methods in
  Natural Language Processing (EMNLP)}, pp.\  4222--4235, Online, November
  2020. Association for Computational Linguistics.
\newblock \doi{10.18653/v1/2020.emnlp-main.346}.
\newblock URL \url{https://aclanthology.org/2020.emnlp-main.346}.

\bibitem[Song et~al.(2020)Song, Rush, and
  Shmatikov]{song-etal-2020-adversarial}
Song, C., Rush, A., and Shmatikov, V.
\newblock Adversarial semantic collisions.
\newblock In \emph{Proceedings of the 2020 Conference on Empirical Methods in
  Natural Language Processing (EMNLP)}, pp.\  4198--4210, Online, November
  2020. Association for Computational Linguistics.
\newblock \doi{10.18653/v1/2020.emnlp-main.344}.
\newblock URL \url{https://aclanthology.org/2020.emnlp-main.344}.

\bibitem[Thomason et~al.(2016)Thomason, Sinapov, Svetlik, Stone, and
  Mooney]{thomason2016learning}
Thomason, J., Sinapov, J., Svetlik, M., Stone, P., and Mooney, R.~J.
\newblock Learning multi-modal grounded linguistic semantics by playing" i
  spy".
\newblock In \emph{IJCAI}, pp.\  3477--3483, 2016.

\bibitem[Wang et~al.(2022)Wang, Mishra, Alipoormolabashi, Kordi, Mirzaei,
  Arunkumar, Ashok, Dhanasekaran, Naik, Stap, et~al.]{wang2022benchmarking}
Wang, Y., Mishra, S., Alipoormolabashi, P., Kordi, Y., Mirzaei, A., Arunkumar,
  A., Ashok, A., Dhanasekaran, A.~S., Naik, A., Stap, D., et~al.
\newblock Benchmarking generalization via in-context instructions on 1,600+
  language tasks.
\newblock \emph{arXiv preprint arXiv:2204.07705}, 2022.

\bibitem[Warstadt et~al.(2018)Warstadt, Singh, and Bowman]{warstadt2018neural}
Warstadt, A., Singh, A., and Bowman, S.~R.
\newblock Neural network acceptability judgments.
\newblock \emph{arXiv preprint arXiv:1805.12471}, 2018.

\bibitem[Wei et~al.(2021)Wei, Bosma, Zhao, Guu, Yu, Lester, Du, Dai, and
  Le]{Wei2021FinetunedLM}
Wei, J., Bosma, M., Zhao, V., Guu, K., Yu, A.~W., Lester, B., Du, N., Dai,
  A.~M., and Le, Q.~V.
\newblock Finetuned language models are zero-shot learners.
\newblock \emph{ArXiv}, abs/2109.01652, 2021.

\bibitem[Williams et~al.(2018)Williams, Nangia, and
  Bowman]{williams-etal-2018-broad}
Williams, A., Nangia, N., and Bowman, S.
\newblock A broad-coverage challenge corpus for sentence understanding through
  inference.
\newblock In \emph{Proceedings of the 2018 Conference of the North {A}merican
  Chapter of the Association for Computational Linguistics: Human Language
  Technologies, Volume 1 (Long Papers)}, pp.\  1112--1122, New Orleans,
  Louisiana, June 2018. Association for Computational Linguistics.
\newblock \doi{10.18653/v1/N18-1101}.
\newblock URL \url{https://aclanthology.org/N18-1101}.

\bibitem[Wortsman et~al.(2021)Wortsman, Ilharco, Li, Kim, Hajishirzi, Farhadi,
  Namkoong, and Schmidt]{wortsman2021robust}
Wortsman, M., Ilharco, G., Li, M., Kim, J.~W., Hajishirzi, H., Farhadi, A.,
  Namkoong, H., and Schmidt, L.
\newblock Robust fine-tuning of zero-shot models.
\newblock \emph{arXiv preprint arXiv:2109.01903}, 2021.

\bibitem[Yin et~al.(2019)Yin, Hay, and Roth]{yin-etal-2019-benchmarking}
Yin, W., Hay, J., and Roth, D.
\newblock Benchmarking zero-shot text classification: Datasets, evaluation and
  entailment approach.
\newblock In \emph{Proceedings of the 2019 Conference on Empirical Methods in
  Natural Language Processing and the 9th International Joint Conference on
  Natural Language Processing (EMNLP-IJCNLP)}, pp.\  3914--3923, Hong Kong,
  China, November 2019. Association for Computational Linguistics.
\newblock \doi{10.18653/v1/D19-1404}.
\newblock URL \url{https://aclanthology.org/D19-1404}.

\bibitem[Zeng \& Wagner(2002)Zeng and Wagner]{zeng2002modeling}
Zeng, X. and Wagner, M.
\newblock Modeling the effects of epidemics on routinely collected data.
\newblock \emph{Journal of the American Medical Informatics Association},
  9\penalty0 (Supplement\_6):\penalty0 S17--S22, 2002.

\bibitem[Zhang et~al.(2019)Zhang, Kishore, Wu, Weinberger, and
  Artzi]{zhang2019bertscore}
Zhang, T., Kishore, V., Wu, F., Weinberger, K.~Q., and Artzi, Y.
\newblock Bertscore: Evaluating text generation with bert.
\newblock \emph{arXiv preprint arXiv:1904.09675}, 2019.

\bibitem[Zhang et~al.(2015)Zhang, Zhao, and LeCun]{10.5555/2969239.2969312}
Zhang, X., Zhao, J., and LeCun, Y.
\newblock Character-level convolutional networks for text classification.
\newblock In \emph{Proceedings of the 28th International Conference on Neural
  Information Processing Systems - Volume 1}, NIPS'15, pp.\  649–657,
  Cambridge, MA, USA, 2015. MIT Press.

\bibitem[Zhong et~al.(2021)Zhong, Lee, Zhang, and
  Klein]{zhong-etal-2021-adapting-language}
Zhong, R., Lee, K., Zhang, Z., and Klein, D.
\newblock Adapting language models for zero-shot learning by meta-tuning on
  dataset and prompt collections.
\newblock In \emph{Findings of the Association for Computational Linguistics:
  EMNLP 2021}, pp.\  2856--2878, Punta Cana, Dominican Republic, November 2021.
  Association for Computational Linguistics.
\newblock URL \url{https://aclanthology.org/2021.findings-emnlp.244}.

\end{thebibliography}
\bibliographystyle{icml2022}

\newpage
\appendix

\onecolumn

\section{Using BERT-score for Evaluation}\label{appendix:use-bert-score}
\begin{figure}
    \centering
    \includegraphics[width=\columnwidth]{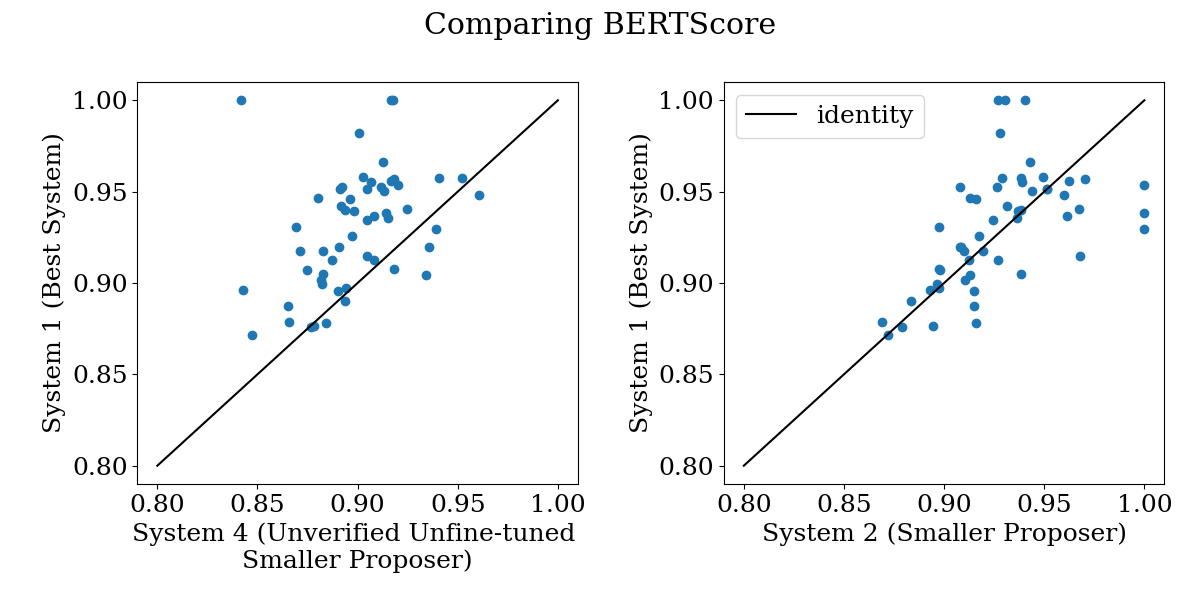}
    \caption{
    We compare System \circled{1} and \circled{4} with BERTScore \cite{zhang2019bertscore} on the left and \circled{1} and \circled{2} on the right.
    Each dot represents a binary task the $y/x$ value is the performance of a system-generated hypothesis evaluated by BERTScore. 
    Our best system \circled{1} is clearly outperforming the worst \circled{4} (left), but the difference between the 1st and the 2nd system becomes hard to tell (right). 
    }
    \label{fig:bertscoresanity}
\end{figure}

We generate scatter plots to compare our best system \circled{1} with the worst system \circled{4} and our second best system \circled{2} in \Cref{fig:bertscoresanity} to double-check that we used the metric correctly.
Despite the the small absolute difference (3\%) in the reported numbers, BERTScore does robustly tell the difference between system system \circled{1} and \circled{4}. 
On the other hand, however, it has trouble discriminating our first and second best system: after squinting at the results hard enough, we find that \circled{1} outperforms \circled{2} by 0.3 points on average; across binary tasks, \circled{1} outperforms \circled{2} more than 0.5 points for 46\% of the time, while \circled{2} outperforms \circled{1} by more than 0.5 points 31\% of the time. 
Therefore, BERTscore does agree that \circled{1} is better than \circled{2}.
Nevertheless, we felt that this metric is not discriminative and interpretable enough, so we had to rely on human evaluation (\Cref{sec:benchmark}). 

\section{Top-K Performance} \label{appendix:top-k}
We calculate the performance of the top-$K$ descriptions by our system according to our manual evaluation, where $K$ ranges from $1$ to $5$.
\Cref{tab:top-k} shows the results. 

\begin{table}[h]
    \centering
\begin{tabular}{llllll}
\toprule
{} &               \circled{1} Best &               \circled{2} Smaller &               \circled{3} No Fine-tune &               \circled{4} No Re-rank &               \circled{5} Memorize \\
\midrule
A &  13/26/28/30/31 &  14/17/21/21/22 &    \phantom{0}6/\phantom{0}8/10/10/11 &       \phantom{0}2/\phantom{0}3/\phantom{0}3/\phantom{0}3/\phantom{0}4 &       \phantom{0}2/\phantom{0}3/\phantom{0}5/\phantom{0}5/\phantom{0}5 \\
B &  19/14/13/11/10 &   10/10/8/10/11 &       \phantom{0}5/\phantom{0}6/\phantom{0}6/\phantom{0}6/\phantom{0}6 &       \phantom{0}1/\phantom{0}0/\phantom{0}0/\phantom{0}0/\phantom{0}0 &       \phantom{0}5/\phantom{0}5/\phantom{0}5/\phantom{0}5/\phantom{0}5 \\
C &      16/\phantom{0}8/\phantom{0}7/\phantom{0}7/\phantom{0}7 &  17/14/12/12/10 &    \phantom{0}7/\phantom{0}9/10/10/10 &       \phantom{0}2/\phantom{0}3/\phantom{0}5/\phantom{0}5/\phantom{0}6 &  16/19/19/21/21 \\
D &       \phantom{0}6/\phantom{0}6/\phantom{0}6/\phantom{0}6/\phantom{0}6 &  13/13/13/11/11 &  36/31/28/28/27 &  49/48/46/46/44 &  31/27/25/23/23 \\
\bottomrule
\end{tabular}
    \caption{
    Similar to \Cref{tab:main-results},
    \circled{1} represents our best system with the largest fine-tuned proposer, \circled{2} with a smaller fine-tuned proposer, \circled{3} without fine-tuning, \circled{4} without re-ranking, and \circled{5} with the memorization proposer. 
    For each task, we choose the top-$K$ descriptions according to the verifier, and find the highest human rating among the top-$K$; we then count how often each rating occurs across 54 binary tasks.
    We report $K$ from 1 to 5 separated by ``/'' in each cell. 
    Notice that only row (A) is guaranteed to increase as $k$ increases, since we are counting the frequency of the highest ranking; e.g., using five rather than one description can change the highest rating from (B) to (A), thus decreasing the count of (B). 
    }
    \label{tab:top-k}
\end{table}

We report the statistical significance of comparing different systems by their best-of-top-5 descriptions.
We first examine how often a system generates a ``A'' level description across 54 binary classification datasets;
as a result \circled{1} $>$ \circled{2} with $p=9.3 \times 10^{-3}$, \circled{2} $>$ \circled{3} with $p=3.2 \times 10^{-3}$, \circled{3} $>$ \circled{4} with $p=2.5 \times 10^{-2}$, and \circled{2} $>$
\circled{5} with $p = 4.2 \times 10^{-5}$.
We next examine how often a system generates a ``A'' or ``B'' level description;
as a result \circled{1} $>$ \circled{2} with $p=9.7 \times 10^{-3}$, \circled{2} $>$ \circled{3} with $p=1.7 \times 10^{-4}$, \circled{3} $>$ \circled{4} with $p=5.4 \times 10^{-4}$, and \circled{2} $>$
\circled{5} with $p = 1.6 \times 10^{-6}$.

\section{Example Positive Samples in \cref{sec:collect}}
We list a few example hypotheses along with positive sample text (i.e., GPT-3 generated texts that are likely to satisfy the hypotheses), indxed with bullet points. 

\noindent\textbf{Hypothesis: } \textit{contains internet emoticons.}
\begin{itemize}
    \item \textit{ :) - I'm happy}
    \item \textit{:) :D :o :(}
    \item \textit{(:-I'm so excited to tell you about my plans for the future. ): I hope they work out!}
\end{itemize}

\noindent\textbf{Hypothesis: }\textit{contains an acronym or abbreviation that might be used online.}
\begin{itemize}
    \item \textit{ SMH is an acronym for shaking my head. It is used when someone reacts to something funny or shocking.}
    \item \textit{OMG. I think it\'s my new BF.}
    \item \textit{IDK - I don't knowTTYL - talk to you later.}
\end{itemize}

\noindent\textbf{Hypothesis: } \textit{is about school life.}
\begin{itemize}
    \item \textit{Entering the courtyard of the student center, I had a feeling that there was something different about today.}
    \item \textit{The first day of school is always nerve-wracking. You're the new kid, and you have to make friends fast.}
    \item \textit{I came to this country in the summer of 2010. I attended a public high school in my new country and was really scared because I did not know what to expect from it.}
\end{itemize}

\noindent\textbf{Hypothesis: } \textit{is about money}.
\begin{itemize}
    \item \textit{It's so easy to think about all of the things that you could do with your life if you had more disposable income, but it's hard work to make more than enough. I'm not sure what my goal is when it comes to my finances, but I know that whatever happens in the future has got to be better than this.}
    \item \textit{ There's a recession going on and cash flow is scarce.}
    \item \textit{Money is very important in today's society. It lets you do what you want to do and nothing can stop you.}
\end{itemize}

\noindent\textbf{Hypothesis: }\textit{describes cooking.}
\begin{itemize}
    \item \textit{If you can perfect a recipe, it's amazing. I love being able to read through a recipe and have the confidence that I have all of the ingredients listed.}
    \item \textit{I love to watch the way that each ingredient has their own value, and how they all come together to make a dish.}
    \item \textit{One rainy night, I decided to make spaghetti.}
\end{itemize}

\section{Example Descriptions and Their Ratings}

For each binary task, we present the human annotation, the best descriptions from the top-5 descriptions by system \circled{1}, and our similarity rating in \Cref{tab:example-description}.
\begin{table}[]
    \centering
    \begin{tabular}{l|l|l}
    Human Annotations & Descriptions by Our System & Rating\\
    \hline
       \textit{is religious} &                                           \textit{is religious} & (A) \\
\textit{is against feminism} &                             \textit{is a criticism of feminism} & (A) \\
\textit{is about math or science} &                                       \textit{is about science} & (B) \\
\textit{asks about a location} &                                  \textit{asks about a location} & (B) \\
\textit{contains a good movie review} &                                       \textit{praises the film} & (A) \\
       \textit{is offensive} &                                 \textit{is a Twitter hate-rant} & (C) \\
\textit{is related to computer science} &            \textit{is a description of a computer-based system} & (B) \\
\textit{is against environmentalist} &                  \textit{is a denial of climate change science} & (C) \\
 \textit{is against Hillary} &                      \textit{is a criticism of Hillary Clinton} & (A) \\
      \textit{is pro-choice} &                          \textit{advocates for abortion rights} & (A) \\
\textit{is about research in statistics} &             \textit{presents a research on a statistical topic} & (A) \\
\textit{is related to infrastructure} &                              \textit{mentions natural disaster} & (D) \\
\textit{is about entertainment} &               \textit{is related to the entertainment industry} & (B) \\
\textit{is environmentalist} &                         \textit{shows an environmental concern} & (A) \\
\textit{is related to health} &                         \textit{is about the topic of ``health''} & (A) \\
     \textit{contains irony} &                                   \textit{is sarcastic in tone} & (A) \\
   \textit{supports hillary} &           \textit{is a positive sentence about Hillary Clinton} & (A) \\
\textit{contains a definition} &                        \textit{is about learning something new} & (B) \\
\textit{is related to terrorism} &                                     \textit{is about terrorism} & (A) \\
\textit{expresses a need for water} &                                \textit{is about water shortage} & (A) \\
     \textit{involves crime} &                                  \textit{is describing clashes} & (C) \\
\textit{is related to sports} &                                        \textit{is about sports} & (A) \\
\textit{is related to a medical situation} &                      \textit{is related to the topic of health} & (B) \\
\textit{describes a situation where people need food} &                \textit{is about the situation of food shortage} & (A) \\
        \textit{is pro-life} &               \textit{can be categorized as a pro-life message} & (A) \\
\textit{contains subjective opinions} &                                 \textit{is a review of a movie} & (D) \\
\textit{asks for an opinion} &                                     \textit{is asking for help} & (D) \\
\textit{is more related to computers or internet} &                                      \textit{is about computer} & (B) \\
\textit{expresses need for utility, energy or sanitation} &                 \textit{contains a word related to electricity} & (C) \\
  \textit{is sports related} &                     \textit{is about a topic related to sports} & (A) \\
  \textit{asks for a number} & \textit{contains a question ...}$^{*}$ & (A) \\
\textit{describes a situation where people need to evacuate} &             \textit{describes a situation involving evacuation} & (A) \\
\textit{is a more objective description of what happened} &                            \textit{is a plot summary of a film} & (D) \\
\textit{is physics research} &                            \textit{is about a physics research} & (A) \\
\textit{is about world news} &                         \textit{is a news article on a country} & (C) \\
\textit{looks more like business news} &                               \textit{deals with economic news} & (A) \\
\textit{describes a situation where people need shelter} &                                    \textit{is about earthquake} & (C) \\
          \textit{is a spam} &                                        \textit{is a ``spam'' SMS} & (A) \\
\textit{contains grammar errors} &                             \textit{is grammatically incorrect} & (A) \\
\textit{asks about an entity} &              \textit{contains a word that rhymes with ``tree''} & (D) \\
\textit{is about math research} &                  \textit{is about a mathematics research paper} & (A) \\
  \textit{supports feminism} &                              \textit{is in support of feminism} & (A) \\
\textit{asks for factual information} &         \textit{is a request for immigration related questions} & (D) \\
  \textit{is more political} &                                      \textit{is about politics} & (A) \\
\textit{is against religion} &            \textit{has a negative connotation towards religion} & (A) \\
\hline
    \end{tabular}
    \caption{For each binary task, we present the human annotation, the best descriptions from the top-5 descriptions by system \circled{1}, and our similarity rating, with (A) being the highest (\Cref{sec:benchmark}).
    \\\hspace{\textwidth}
    $^{*}$: ``\textit{contains a question that can be answered with a number}''; truncated from the column to save sapce.
    }
    \label{tab:example-description}
\end{table}

\section{A Unifying View} \label{appendix:method-remark}

\begin{figure}[h]
    \centering
    \includegraphics[width=0.7\columnwidth]{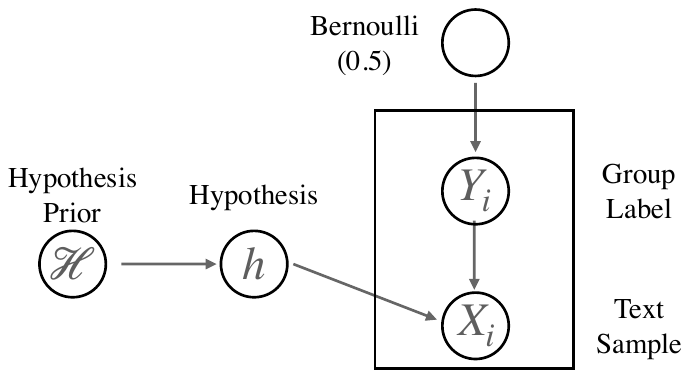}
    \caption{A unifying graphical model interpretation of our framework, where the verifier, the proposer, and the conditional generator can be all written as posterior estimators.}
    \label{fig:graphical}
\end{figure}

We present a unifying graphical model for the hypothesis $h$, the samples $X_{1\dots K}$, and the group labels $Y_{1\dots K}$ (\Cref{fig:graphical}), where $Y_{i} \in \{0, 1\}$ indicating whether $X_{i}$ is from distribution $D_{0}$ or $D_{1}$.
Although we did not implement it in our paper, we find it helpful as a mental model to generate future research directions.
The graphical model factorizes as:
\begin{equation}
    p(h, X_{1\dots K}, Y_{1\dots K}) = p(h) \prod_{i=1}^{K} p(X_{i}|Y_{i}, h)P(Y_{i}).
\end{equation}
Under this framework, the goal of generating a natural language hypothesis becomes posterior estimation: 
 \begin{equation} \label{eq:bayes-target-problem}
     p(h|X_{1\dots K}, Y_{1\dots K}) \propto p(h) \prod_{i=1}^{K} p(Y_{i}|X_{i}, h).
\end{equation}

The verifier can also be written as $\hat{p}(Y|X, h)$, the proposer as $\hat{p}(h|X_{1\dots 5}, Y_{1\dots 5})$, the conditional generator as $\hat{p}(X|Y, h)$, and the hypothesis space as a prior $\hat{p}(h)$,\footnote{which our paper defines through manual curation of the hypothesis and modelled as a uniform distribution during inference.} all of which can be directly approximated by a fine-tuned language model. 
To fine-tune these approximators, it suffices to obtain the complete data $h, X_{*}$, and $Y_{*}$.
Our work only fine-tuned the verifier and the proposer, but the conditional generator $\hat{p}(X|Y, h)$ and $\hat{p}(h)$ can also be fine-tuned.
We only supervised $\hat{p}$ through querying human about $p(Y|X, h)$, but other forms of queries are also possible.
Finally, it is not necessary to follow the recipe in our paper to generate the complete data: we could alternatively first generate $X$ and $h$, and then generate $Y$ accordingly.
Human supervision is also not strictly necessary to generate the complete data: we can purely sample data from some approximators to fine-tune other ones, thus achieving self-supervision through cycle consistency.

\section{Original Sources of the Binary Tasks}
The 54 binary tasks are from \citet{maas-EtAl:2011:ACL-HLT2011}, \citet{yin-etal-2019-benchmarking}, \citet{barbieri-etal-2020-tweeteval}, \citet{10.5555/2969239.2969312}, \citet{yin-etal-2019-benchmarking}, 
\citet{warstadt2018neural}, 
\citet{almeida2013towards}, 
\citet{pang-lee-2004-sentimental}, \citet{li2002learning},
\citet{mihaylova-etal-2019-semeval},
and an abstract classification dataset\footnote{\url{https://www.kaggle.com/abisheksudarshan/ topic-modeling-for-research-articles? select=Train.csv}}. 

\section{Notes on Code and Model Release} \label{appendix:code}
We release our code and data with the following link \url{https://github.com/ruiqi-zhong/DescribeDistributionalDifferences}. 

We cannot directly share our GPT-3 based proposer, since it has to be accessed through the OpenAI API using our own key. 
To make it easier for other researchers to use our system, we trained another proposer by fine-tuning T5 \cite{raffel2019exploring} on a mixture of 1) our collected data, and 2) a large dataset  \citet{wang2022benchmarking} to learn to follow task instructions.
Additionally, we implemented the ensembling approach mentioned in \cref{sec:disussion}. 
Though we have not rigorously benchmarked the new proposer, it seems to be roughly comparable to the proposer based on GPT-3 Davinci (175B parameters), and it can be openly shared, downloaded, and run locally. 

\end{document}